\pdfoutput=1

\documentclass{article}

\usepackage{newunicodechar}

\usepackage{natbib}
\usepackage[main,final,nonatbib]{latex/neurips_2026}

\usepackage{latexsym}

\usepackage[T1]{fontenc}

\usepackage[utf8]{inputenc}

\usepackage{microtype}

\usepackage{inconsolata}

\usepackage{graphicx}
\usepackage{placeins}
\usepackage{booktabs}
\usepackage{multirow}
\usepackage{array}
\usepackage{caption}
\usepackage{ragged2e}
\PassOptionsToPackage{dvipsnames,table}{xcolor}
\usepackage{xcolor}
\usepackage[most]{tcolorbox}
\usepackage{setspace}

\usepackage{textcomp, gensymb}
\usepackage{soul}
\soulregister\ref7
\soulregister\pageref7
\soulregister\cite7
\soulregister\S7
\usepackage{amsmath}
\usepackage{subcaption}
\usepackage{enumitem}
\usepackage{tabularx,seqsplit}
\newcolumntype{Y}{>{\raggedright\arraybackslash}X}

\definecolor{rowgray}{gray}{0.95} 

\usepackage{amssymb}
\usepackage{pifont}
\usepackage{algorithm}
\usepackage{algpseudocode}
\usepackage{hyperref}

\title{ESCROW: Guarded and Dual-Objective Continual Maintenance for Agents in Policy-Governed Enterprise Workflows}

\author{
  \textbf{Ruoqi Shu}\thanks{Equal contribution.}, 
  \textbf{Chen Dan}\footnotemark[1], 
  \textbf{Xuhui Wang}\footnotemark[1], 
  \textbf{Tianhua Xu}\footnotemark[1]\\
  \textbf{Mengxi Luo}, 
  \textbf{Yanming Mai}, 
  \textbf{Bo Wan}\\
  BMO Financial Group\\
    \texttt{\{%
    \href{mailto:ruoqi.shu@bmo.com}{ruoqi.shu},
    \href{mailto:chendaisy.dan@bmo.com}{chendaisy.dan},
    \href{mailto:xuhuieric.wang@bmo.com}{xuhuieric.wang},
    \href{mailto:TianhuaMatthew.Xu@bmo.com}{TianhuaMatthew.Xu},}\\
    \texttt{%
    \href{mailto:Mengxi.Luo@bmo.com}{Mengxi.Luo},
    \href{mailto:yanming.Mai@bmo.com}{yanming.mai},
    \href{mailto:bo.wan@bmo.com}{bo.wan}\}@bmo.com}
}

\begin{document}
\maketitle

\begin{abstract}
LLM agents increasingly run policy-bound enterprise workflows, where they must apply rules consistently and stay auditable. Deploying such an agent is the start of its long-term maintenance cycle: it must adapt to a stream of \emph{operational signals}, yet reliably turning these sparse, unlabeled signals into reusable skill revisions is hard, and a careless update can trade one task category's accuracy for the overall gain, revive a resolved failure, or land at an undeployable cost. We present \textbf{ESCROW}, a post-deployment maintenance framework that updates an agent's external, reviewable skills under a \emph{Strict Update Boundary}: the LLM proposes candidate revisions, but only an empirically evaluated version is deployed. It combines distributed diagnosis with consensus, a per-category non-regression guard, cross-cycle anti-regression, and accuracy--cost Pareto search, emitting a versioned, auditable diff per change. In real production on our internal financial document-auditing system, it attains the strongest evaluated accuracy--cost trade-off among baselines, with a transfer probe on public $\tau$-bench.
\end{abstract}

\section{Introduction}
Large language models now power agents that plan, call tools, and follow explicit procedural skills to automate business workflows~\citep{wei2022chain, yao2023react, wang2023voyager, skillsbench}, and such agents already run policy-bound work such as financial auditing, customer service, and public administration~\citep{shu2025lava, zwerdling2025policy}.

Deploying such an agent is not the end of the work but the start of its maintenance: it must keep up with a stream of \emph{operational signals}---amended policies, drifting documents, reviewer notes, overrides, and failure logs---that production systems already emit. Updating the agent's external procedural skills from these signals is a reviewable alternative to retraining the model, but doing so reliably under regulation is hard~\citep{yao2024tau}. \emph{(i)}~Signals arrive scattered and under-specified: no single one specifies a reusable fix, and merging many into one coherent revision is error-prone. \emph{(ii)}~An update that raises overall accuracy can quietly make one kind of task worse, and the overall number hides it. \emph{(iii)}~As new signals keep arriving, skills are rewritten again and again, so a later update can revive a previously resolved failure. \emph{(iv)}~The most accurate candidate is not always deployable once cost and auditability are weighed. Existing methods for improving deployed agents~\citep{evoskill, autoskill, skillrl} do not target these requirements.

\begin{figure}[t]
\centering
\includegraphics[width=0.475\linewidth]{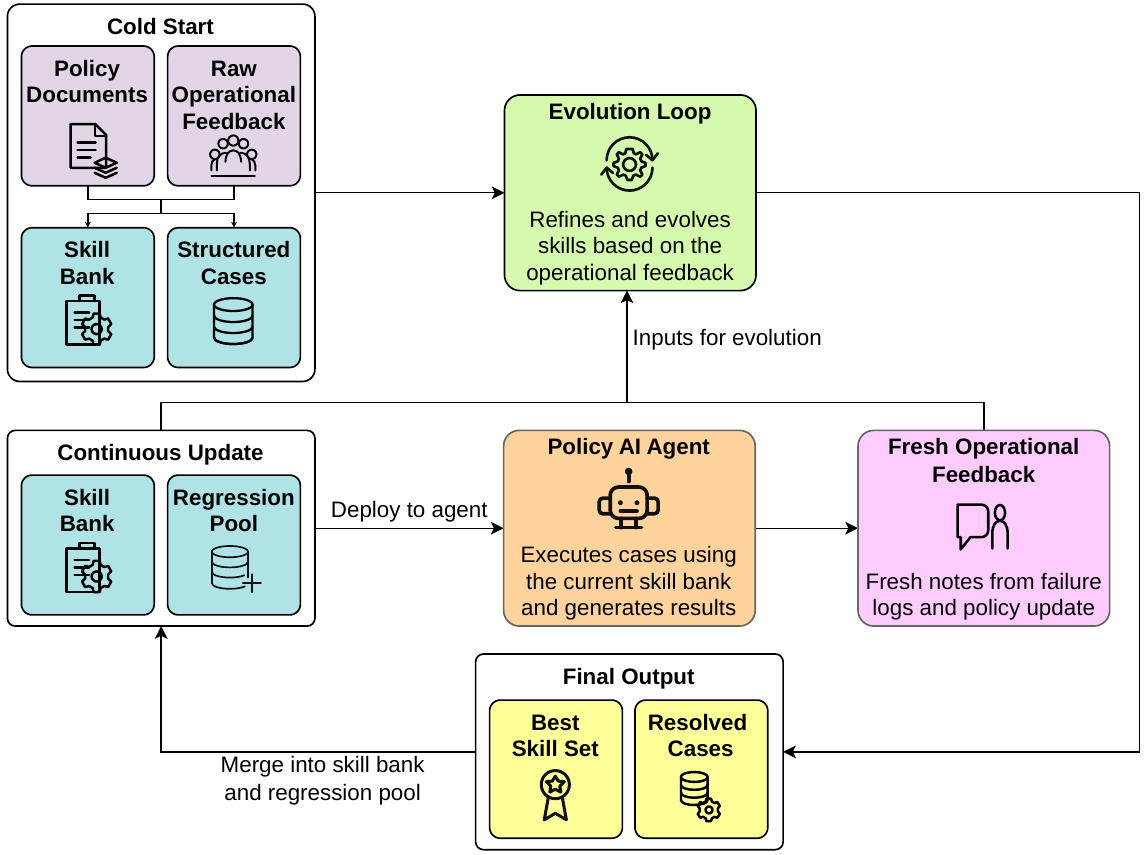}
\caption{Overview of the ESCROW pipeline.
A one-time \emph{cold start} (\S\ref{sec:cold_start}) turns the policy corpus and any raw operational feedback into a skill bank and a set of structured cases.
The \emph{evolution loop} (\S\ref{sec:retrieval}--\S\ref{sec:frontier}) then evolves the skills implicated by the current cases into a new skill set, whose improved skills and resolved cases are merged back into the skill bank and a persistent \emph{regression pool}.
The updated bank is redeployed to the agent; subsequent runs fire periodically or when policies change, with the bank and pool accumulating across runs.}
\label{fig:overview}
\end{figure}

We introduce \textbf{ESCROW}, a post-deployment maintenance framework for policy-governed enterprise agents that learns skill revisions from operational signals under a \emph{Strict Update Boundary}: the LLM proposes candidate revisions, but only an empirically evaluated version is deployed.
It runs in two stages (Figure \ref{fig:overview}): a one-time \emph{cold start} turns existing enterprise assets into a deployable skill bank and an evaluation benchmark, even before any signals arrive; thereafter an \emph{evolution loop} runs periodically, re-improving the responsible skills as new signals accumulate and redeploying them automatically. Because ESCROW maintains external skills rather than the agent's interface, it applies whether the agent runs autonomously or through dialogue.

Our contributions are as follows.
\textbf{Framing.} We cast continuous improvement of a deployed agent as \emph{post-deployment maintenance} of external, reviewable skills, organized around three failure modes \emph{(i)--(iii)} and one deployment tension \emph{(iv)}.
\textbf{Mechanisms.} Under the Strict Update Boundary, ESCROW combines four designs: \emph{distributed diagnosis with consensus} localizes reusable fixes from operational signals; a \emph{per-category non-regression guard} anchors every category before an update is admitted; a \emph{cross-cycle anti-regression mechanism} rechecks candidates against previously fixed cases; and an \emph{accuracy--cost Pareto} search exposes deployable operating points---to our knowledge, a combination no prior frozen-model method assembles for regulated maintenance.
\textbf{Deployment.} ESCROW cold-starts from policies alone, emits a versioned, auditable natural-language diff for every change, and plugs into existing agent systems without a rebuild.

We deploy and evaluate ESCROW on our internal financial document-auditing system (\textbf{FinDAS}) in real production, where it yields the strongest evaluated accuracy--cost trade-off among baselines, with a transfer probe on the public $\tau$-bench.


\section{Related work}
Keeping a deployed, policy-governed agent correct is an ongoing maintenance
problem rather than a one-time fit.
Systems built for this setting often constrain behaviour rather than change
it---validating documents against symbolic rules \citep{shu2025lava}, compiling
policy into runtime guards \citep{zwerdling2025policy}, or deferring to human
oversight \citep{he2025selfimprove}---while those that route production signals
back into improvement stop at collecting them \citep{zhao2025aitl}, though
industry practice argues the loop should close \citep{feizi2026relai}. The
procedures the agent actually follows stay static.

Work that does rewrite those procedures sits outside deployment. Frozen-model
methods refine skill files from execution traces and failures
\citep{skillsbench, evoskill, autoskill, trace2skill, skillx, mementoskills,
asda}; others make skills scaffolding for reinforcement learning
\citep{skillrl, metaclaw, skillzero, wang2026sage}, yielding weights costly to
audit or roll back. GEPA~\citep{agrawal2026gepareflectivepromptevolution}
and GRASP~\citep{moll2026graspgatedregressionawareskill} add reflective mutation
and regression-aware gating, and a recent
survey~\citep{xu2026agentskillslargelanguage} flags skill reuse at scale as
open.

ESCROW keeps that substrate---a frozen model, skills in natural language---but
not that setting. Under maintenance, updates land repeatedly in a live system,
where a category floor cannot be traded for an aggregate gain, an earlier fix
must survive later rounds, and cost decides deployability. This is the gap
ESCROW occupies: post-deployment maintenance of policy-governed enterprise
agents through external, reviewable skills, admitted on evidence rather than on
proposal.

\section{Method}
\subsection{Problem formulation}
\label{sec:problem}
ESCROW is a post-deployment maintenance framework that separates LLM-based skill revision from evidence-based deployment decisions. We formalize maintenance as a dual-objective optimization over the skill bank $\mathcal{S} = \{s_1, \ldots, s_L\}$ that backs the agent at serving time, encoding per-category reliability, cost-awareness, and non-regression across updates.
Because the underlying model is fixed, the correctness and cost of executing real business cases are mainly determined by $\mathcal{S}$. 
We evaluate $\mathcal{S}$ on a set of cases, running each $n$ times to absorb LLM stochasticity, which yields pass rate $\text{PR}(\mathcal{S})$ scored with the unbiased pass@$k$ estimator~\citep{chen2021codex} and cost $\text{Cost}(\mathcal{S})$ as the average LLM token consumption per case. Cases are partitioned into operational categories $\mathcal{C}$ that reflect heterogeneous business risks (e.g., routine validation vs.\ high-impact escalation), and pass rate is reported per category as $\text{PR}_c(\mathcal{S})$ for $c \in \mathcal{C}$.
ESCROW jointly maximizes pass rate and minimizes cost, subject to a \emph{per-category non-regression} constraint:
\begin{equation}
\label{eq:objective}
\begin{aligned}
\max_{\mathcal{S}} \;\; & \big(\text{PR}(\mathcal{S}),\; -\text{Cost}(\mathcal{S})\big) \\
\text{s.t.} \;\; & \text{PR}_c(\mathcal{S}) \geq \text{PR}_c(\mathcal{S}_{\text{init}})\;\; \forall c \in \mathcal{C}
\end{aligned}
\end{equation}
where $\mathcal{S}_{\text{init}}$ is the skill bank at the start of maintenance.
Because pass rate and cost can conflict, ESCROW optimizes in the Pareto sense during maintenance (\S\ref{sec:frontier}).
ESCROW is run periodically after deployment, forming a maintenance loop rather than a one-off optimization. Between runs, it ingests new operational signals into the case set, so each run adapts $\mathcal{S}$ to conditions that have shifted since the last. A run edits a working scope $\mathcal{P} \subseteq \mathcal{S}$ and commits the improved $\mathcal{P}^*$ back, so improvements accumulate across runs instead of restarting. In parallel, a \emph{persistent regression pool} absorbs the cases each run resolves and is re-checked on later updates, guarding against revisions that would silently reintroduce previously resolved failures.

\begin{figure}[t]
\centering
\includegraphics[width=1\linewidth]{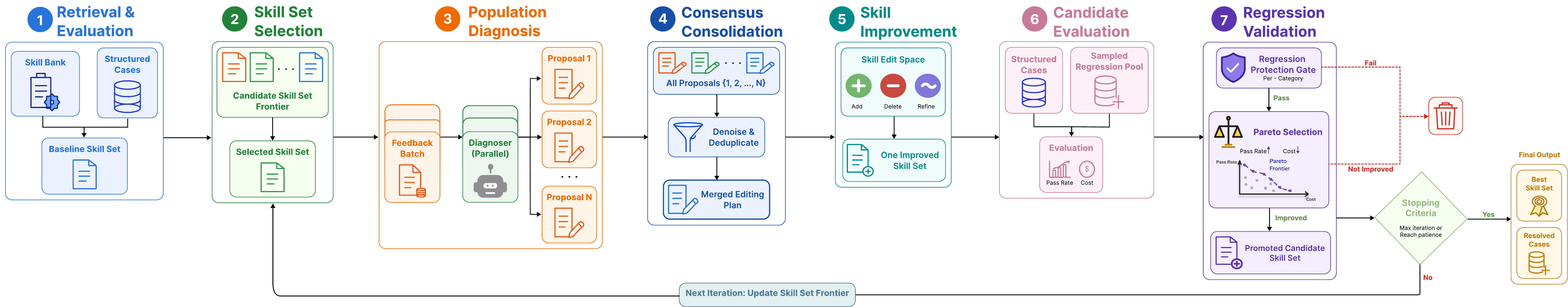}
\caption{The evolution loop of ESCROW, which searches a Pareto frontier of skill sets trading off pass rate and cost.
A run first scores the retrieved $\mathcal{P}_0$ to fix a baseline $\mathcal{E}_0$ that anchors the gate and seeds the frontier.
Each iteration then draws a parent from the frontier and lets a population of diagnosers analyze feedback against it in parallel; their proposals are consolidated into a single edit plan that rewrites the parent into a child.
The child must first clear a per-category non-regression gate (Eq.~\ref{eq:gate}), evaluated on current cases and a sampled subset of the regression pool, and then survive Pareto dominance (Eq.~\ref{eq:dominance}) to enter the frontier.
Once stopping criteria are met, a frontier member $\mathcal{P}^*$ is selected and committed back to the skill bank, and its resolved cases join the regression pool.}
\label{fig:architecture}
\end{figure}

\begin{algorithm}[t]
\small
\caption{ESCROW evolution run}\label{alg:escrow}
\begin{algorithmic}[1]
\Require Skill bank $\mathcal{S}$, structured cases $\mathcal{F}$, regression pool $\mathcal{R}$, iteration budget $T$, population $N$, patience $\tau$
\State $\mathcal{P}_0 \gets \textsc{Retrieve}(\mathcal{S}, \mathcal{F})$ \Comment{Working scope from $\mathcal{S}$}
\State $\mathcal{E}_0 \gets \textsc{Evaluate}(\mathcal{P}_0, \mathcal{F} \cup \textsc{Sample}(\mathcal{R}))$ \Comment{Anchor non-regression}
\State $\textit{frontier} \gets \{(\mathcal{P}_0, \mathcal{E}_0)\}$
\For{$t = 1, \ldots, T$}
    \State $\mathcal{P} \gets \textsc{SelectParent}(\textit{frontier}, t)$ \Comment{Non-greedy}
    \State $\{\mathcal{B}_i\}_{i=1}^N \gets \textsc{Partition}(\mathcal{F}, N, t)$ \Comment{Reshuffled per iteration}
    \State $\{\delta_i\}_{i=1}^N \gets \textsc{Diagnose}(\mathcal{P}, \{\mathcal{B}_i\}_{i=1}^N)$ \Comment{Parallel}
    \State $\hat{\delta} \gets \textsc{Consolidate}(\{\delta_i\}_{i=1}^N)$ \Comment{Consensus}
    \State $\mathcal{P}' \gets \textsc{Rewrite}(\mathcal{P}, \hat{\delta})$
    \State $\mathcal{E}' \gets \textsc{Evaluate}(\mathcal{P}', \mathcal{F} \cup \textsc{Sample}(\mathcal{R}))$
    \If{$\textsc{Gate}(\mathcal{E}', \mathcal{E}_0)$}
        \State $\textit{frontier} \gets \textsc{ParetoUpdate}(\textit{frontier}, \mathcal{P}', \mathcal{E}')$
    \EndIf
    \If{no improvement for $\tau$ iterations} \textbf{break} \EndIf
\EndFor
\State $\mathcal{P}^* \gets \textsc{SelectFromFrontier}(\textit{frontier})$
\State $\mathcal{R}^* \gets \{f \in \mathcal{F} : \mathcal{P}^*\text{ passes }f\}$ \Comment{Cases resolved by $\mathcal{P}^*$}
\State \Return $\mathcal{P}^*, \mathcal{R}^*$
\end{algorithmic}
\end{algorithm}

\subsection{Cold start from enterprise assets}\label{sec:cold_start}

The agent runs on a skill bank $\mathcal{S}$---the target that maintenance optimizes---and is improved against structured cases $\mathcal{F}$ that turn scattered operational signals into a diagnosable, scorable form. ESCROW derives both from existing enterprise assets at cold start; $\mathcal{S}$ is then maintained by the evolution loop, while $\mathcal{F}$ is continually replenished as the deployment runs.

\paragraph{Skill bank construction.}
An LLM splits policy documents $\Pi$ along procedural boundaries and rewrites each segment into a self-contained skill $s_l$ aligned with one verifiable business activity, yielding a skill bank $\mathcal{S}=\{s_1,\dots,s_L\}$ (prompts in \S\ref{app:prompt-splitter}). The bank is \emph{flat-stored, tag-indexed, and reference-composed}: skills sit at one level, are indexed by free-form tags for soft grouping, and express cross-skill dependencies as references. This avoids the cascading re-review that hierarchical edits impose under regulated change management. Each $s_l$ further carries a description $d_l$ (the field retrieval scores against, \S\ref{sec:retrieval}) and a back-link to its authoritative clause in $\Pi$, making any agent action traceable to source policy.

\paragraph{Case construction.}
Every case $f_i = (m_i, t_i)$ couples a \emph{diagnostic record} $m_i$---a natural-language description of the targeted failure mode together with any annotation of root cause---with an \emph{evaluation task} $t_i$, an executable scenario whose rubric encodes the ground-truth outcome. ESCROW first synthesizes a baseline set of such cases from $\Pi$: an LLM identifies policy clauses whose expected behaviour is not yet exercised by $\mathcal{S}$ and drafts a case for each (prompt in \S\ref{app:prompt-surfacer}), so the system can start even before any feedback arrives. When raw operational feedback $\mathcal{F}_{\text{raw}} \neq \emptyset$, an LLM classifier splits it by executability: the reproducible remainder is structured into additional cases of the same $(m_i, t_i)$ form and merged with the synthetic baseline to yield $\mathcal{F} = \{f_i\}$, while \emph{experiential knowledge} $\mathcal{K}$ (advice without a reproducible setup) is instead attached to the relevant skill as read-only supplementary context (details in \S\ref{app:prompt-classifier}).

\paragraph{Continuous feedback ingestion.}
The triage above is a one-time event at system introduction, sorting through the mixed backlog of cases and free-form lessons that legacy systems leave behind. Once this backlog is exhausted, $\mathcal{F}$ is replenished from live operation: between successive evolution runs, new cases are ingested from three channels---\emph{human review} (reviewer notes, overrides, annotations on agent decisions); \emph{automated trajectory analysis} (inspection of agent execution traces); and \emph{policy updates} (amendments to $\Pi$ diffed against $\mathcal{S}$ to surface uncovered clauses). Each channel has its specific LLM-driven converter that refines its raw artifacts into cases of the same $(m_i, t_i)$ form (see prompt in \S\ref{app:prompt-converter} and \S\ref{app:structuredcase} for a worked example).

\subsection{Skill retrieval and baseline anchoring}\label{sec:retrieval}

We now detail the evolution loop (Figure~\ref{fig:architecture}; Algorithm~\ref{alg:escrow}). A run opens by locating the skills responsible for the current cases: each case's diagnostic record $m_i$ is a retrieval query against skill descriptions $\{d_l\}$ via hybrid semantic--lexical scoring (BM25), yielding a working skill set $\mathcal{P}_0 \subseteq \mathcal{S}$.
Skills are ranked by aggregated relevance across all cases; the top-$k$ form $\mathcal{P}_0$, frozen for the run so cross-iteration comparisons reflect content quality rather than scope drift. $\mathcal{P}_0$ is scored on $\{t_i\}$ to obtain baseline $\mathcal{E}_0 = (\text{PR}(\mathcal{P}_0),\, \text{Cost}(\mathcal{P}_0))$, which anchors the non-regression gate for admitting subsequent children (\S\ref{sec:gate}) and seeds the Pareto frontier for retaining non-dominated candidates across iterations (\S\ref{sec:frontier}), as $\{(\mathcal{P}_0, \mathcal{E}_0)\}$.

\subsection{Guarded mutation}\label{sec:propose}

A single rewrite from scattered feedback tends to be narrow or conflicting; ESCROW instead localizes reusable fixes by \emph{distributed diagnosis with consensus}. Each evolution iteration produces a child $\mathcal{P}'$ in four steps: it picks which frontier member to build on (parent selection), diagnoses where that set fails (distributed diagnosis), settles on what to change (consolidation), and rewrites the targeted skills while guarding what already works (skill improvement).

\textbf{Parent selection.} A parent skill set $\mathcal{P}$ is drawn round-robin over current frontier members; this non-greedy schedule keeps the search from collapsing onto one end of the accuracy--cost trade-off.
\textbf{Distributed diagnosis.} ESCROW partitions the diagnostic records $\{m_i\}$ into $N$ disjoint batches $\mathcal{B}_1, \ldots, \mathcal{B}_N$, reshuffled every iteration, and runs one LLM-based diagnoser per batch. Each diagnoser sees the full parent $\mathcal{P}$ plus its $\mathcal{B}_i$, localizes skill-level root causes, and emits a proposal $\delta_i = (\textit{targets}_i, \textit{root\_cause}_i, \textit{plans}_i)$. Batching diversifies diagnosis along two axes. \emph{Across diagnosers}, the disjoint $\{\mathcal{B}_i\}$ give independent diagnoses that consolidation then aggregates, akin to bagging. \emph{Across iterations}, reshuffling lets correlated failures co-occur in fresh combinations, exposing root causes a fixed partition would hide.
\textbf{Consolidation.} A consolidator first applies a vote filter, then makes one LLM call to merge the retained proposals into a unified plan $\hat{\delta}$. The three modes trade breadth for precision: \texttt{none} passes all proposals through; the default \textit{per-skill voting} (\texttt{section}) drops any skill edit proposed by fewer than $q$ diagnosers; and \texttt{cluster} additionally requires agreement on the exact target set.
\textbf{Skill improvement.} For each targeted $s_j \in \mathcal{P}$, an LLM rewrites the procedure under $\hat{\delta}$ to obtain $s_j'$. The rewrite prompt exposes the target's full SKILL.md, sibling skills as one-line summaries, and its auxiliary file inventory, preventing duplication and boundary violations; it also injects a small set of \emph{anchor cases}---previously resolved cases from the regression pool $\mathcal{R}$ whose resolution depends on $s_j$---as behavioural constraints that help preserve behavior the skill previously handled correctly (\S\ref{sec:gate}, \emph{cross-cycle anti-regression}). Untouched skills carry over: $\mathcal{P}' = (\mathcal{P} \setminus \{s_j\}_{j \in \hat{\delta}}) \cup \{s_j'\}_{j \in \hat{\delta}}$.
Because the edit unit is a natural-language procedure, every change is a readable diff a compliance officer can approve and is versioned, leaving a full audit trail. The resulting child $\mathcal{P}'$ then proceeds to scoring.

\subsection{Scoring and admission}\label{sec:gate}
ESCROW evaluates $\mathcal{P}'$ on $\{t_i\}$ plus a sample of the regression pool sized as a fixed fraction of the current cases (\S\ref{app:coverage-growth}), with an agent blind to $\{m_i\}$. Admission is governed by a \emph{per-category non-regression guard}, a safety filter rather than a competition: it asks only whether the child preserves a hard quality floor anchored at $\mathcal{P}_0$. The child clears the gate if
\begin{equation}\label{eq:gate}
\text{PR}_c(\mathcal{P}') \;\geq\; (1 - \epsilon)\,\text{PR}_c(\mathcal{P}_0) \quad \forall\, c \in \mathcal{C},
\end{equation}
where $\epsilon$ is the regression tolerance that relaxes the constraint of Eq.~\ref{eq:objective}, giving the search more intermediate branching headroom than a strict gate ($\epsilon = 0$) would allow.
The bound is per-category to defeat the masking effect: a flat total can hide a regression in one category offset by a gain in another, and in regulated workflows, categories carry asymmetric business risk. Anchoring at $\mathcal{P}_0$ rather than the running parent matters because the latter would let the worst-case floor decay as $(1-\epsilon)^t$ over $t$ iterations, eroding the gate's protection within a run.

\textbf{Cross-cycle anti-regression.} The floor so far guards a single run; maintenance, however, runs many times, and a fix earned in one run can be silently undone by a later one---regression across runs, which one-shot optimizers keep no memory to catch. The same gate handles this because $\mathcal{R}$ persists and accumulates the cases every run resolves (\S\ref{sec:frontier}). Past-resolved cases from $\mathcal{R}$ enter maintenance twice over: the rewrite draws on them as anchor constraints (\S\ref{sec:propose}), conditioning each edit on behaviour earlier runs got right, and scoring folds a fresh sample of them into the evaluation set, where they are scored under the per-category floor together with the current cases. A child that has regressed on this shared history then fails the gate. However, since the sample stays a fixed fraction of the task, its coverage of a growing $\mathcal{R}$ would thin over successive runs; we address this with a \emph{skill-targeted} sampling design whose regression-detection capability depends on skill bank expansion rather than on $|\mathcal{R}|$, so its coverage of a growing pool degrades negligibly (\S\ref{app:coverage-growth}).

\subsection{Dual-objective frontier}\label{sec:frontier}

A gate-clearing child does not enter the frontier automatically; it must win an \emph{accuracy--cost Pareto} competition: because accuracy and cost trade off, production needs a range of operating points---high-value cases justify expensive skill sets while bulk screening needs cheap ones. ESCROW therefore maintains a frontier of skill sets that are non-dominated under
\begin{equation}\label{eq:dominance}
a \succ b \iff \mathbf{o}(a) \geq \mathbf{o}(b) \,\land\, \mathbf{o}(a) \neq \mathbf{o}(b),
\end{equation}
where $\mathbf{o} = (\text{PR}, -\text{Cost})$ is the objective vector, $\geq$ is componentwise. Each iteration, a gate-clearing child enters the frontier only if no incumbent dominates it; any members it dominates are then removed, and the frontier is capped at a fixed size. 

Evolution stops when no child has been \emph{promoted} for $\tau$ consecutive iterations. When several non-dominated points remain, the frontier serves as a \emph{deployment menu} spanning the accuracy--cost trade-off, from which operators select the operating point matching a workload's risk and budget; when a single point dominates, the frontier simply returns it as the operating choice. Either way, the chosen $\mathcal{P}^*$ is committed back to $\mathcal{S}$ in place of $\mathcal{P}_0$, and its resolved cases are merged into $\mathcal{R}$.

\section{Experiments}
\label{sec:experiment}
\subsection{Experimental setup}
\label{sec:setup}

\paragraph{Datasets and environments.}
ESCROW is evaluated on FinDAS, a production system that audits the financial documents---i.e., income documents, tax forms, and financial statements---submitted with mortgage applications.
The evaluation output of each case is a verdict with supporting evidence.
Public document-QA benchmarks~\citep{mathew2021docvqa,opsahlong2026officeqa,chen2021finqa}
pose static questions with fixed per-document gold answers over relatively clean documents;
there, feedback from a wrong answer is a local correction to that case, with no global
policy to reuse. By contrast, FinDAS spans heterogeneous, noisy enterprise-level documents and
applies cross-cutting policies across all of them. Rules derived from these policies chain fields
across pages with conditional, cross-field, and arithmetic checks, making
the policy agent prone to missing edge cases under ambiguous wording and hallucinating
ungrounded values. In deployment, experts flag a wrong case and its
expected output, not the reusable rule it implies---which must generalize without
breaking previously-correct decisions.

Cases from the FinDAS production workflow fall into three categories: 
general covers routine cases with both correct outputs and minor errors; 
hallucination refers to responses not grounded in source documents; 
and special captures edge cases not explicitly covered by policy (\S\ref{app:feedback-samples}).
We sample two disjoint sets of 210 cases each (70 per category), so that each category supplies an equal learning signal:
an evolution set and a held-out test set (\S\ref{app:dataset}).
For protocol consistency, all experiments run inside the $\tau$-bench harness~\citep{yao2024tau},
into which we port FinDAS as a new domain; the cases, policies, and skill bank are the
production ones, and each evolved bank is redeployed in FinDAS. We additionally evaluate its
original retail and airline domains as a transfer probe under a similar setup (\S\ref{app:taubench}).
Agent execution and evaluation are both built on DeepAgents~\citep{deepagents2024};

\paragraph{Default configuration.}
All ESCROW modules run on Claude Sonnet 4.6~\citep{claude-sonnet-4-6} with thinking
enabled, including the policy agent for candidate evaluation, which is the same one used in FinDAS.
ESCROW's full default settings and hyperparameters are listed in \S\ref{app:config}.

\paragraph{Metrics.}
We instantiate $\text{PR}$ as average pass@$k$, reported over all cases
with $95\%$ bootstrap confidence intervals ($B{=}10{,}000$) and per category ($\text{PR}_c$), 
and $\text{Cost}$ as average inference cost in tokens over all cases and all $k$ runs (Eq.~\ref{eq:objective}). 
Every baseline and ESCROW pay this inference cost at deployment; 
only evolution-loop methods pay an additional one-time evolve cost, 
reported in ablation (\S\ref{sec:ablation}).

\subsection{Accuracy and cost across methods}
\label{sec:exp_comparison}
\newcommand{\dgain}[1]{{\scriptsize\textcolor{green!55!black}{$+#1$}}}
\newcommand{\dloss}[1]{{\scriptsize\textcolor{red}{$-#1$}}}
\newcommand{\dzero}{{\scriptsize\textcolor{gray}{$\pm.000$}}}
\newcommand{\ci}[1]{{\scriptsize\textcolor{gray}{[#1]}}}
\label{sec:main}


\begin{table}[t]
\centering
\caption{Main comparison on the held-out FinDAS test set (pass@$k$, $k{=}3$).
\emph{Total} is the average across the three equally sized categories.
Inference cost is in thousands of tokens ($k$), input\,\textbar\,output. \textbf{Bold} marks our method.}
\label{tab:main}
\footnotesize
\renewcommand{\arraystretch}{0.95}
\setlength{\tabcolsep}{3pt}
\begin{tabular}{@{}l cccc c@{}}
\toprule
& \multicolumn{4}{c}{Quality (avg.\ pass@$k$)} & Inf.\ Cost \\
\cmidrule(lr){2-5}\cmidrule(lr){6-6}
Method & General$\uparrow$ & Hallucination$\uparrow$ & Special$\uparrow$ & Total$\uparrow$ & k tok / case$\downarrow$ \\
\midrule
B1\; Raw LLM                   & 0.586 & 0.043 & 0.000 & 0.210 \,\ci{.16,.27} & 375.627 \textbar\ 4.284 \\
B2\; LLM skills                & 0.657 & 0.043 & 0.000 & 0.233 \,\ci{.18,.29} & 341.472 \textbar\ 4.216 \\
B3\; Human skills              & 0.714 & 0.657 & 0.429 & 0.600\,\ci{.53,.67} & 412.338 \textbar\ 4.413 \\
B4\; Batch update              & 0.743 & 0.714 & 0.514 & 0.657\,\ci{.59,.72} & 332.149 \textbar\ 4.231 \\
B5\; AutoSkill                 & 0.786 & 0.800 & 0.543 & 0.710 \,\ci{.65,.77} & 337.823 \textbar\ 6.477 \\
\textbf{ESCROW (ours)} & \textbf{0.857} & \textbf{0.914} & \textbf{0.629} & \textbf{0.800}\,\ci{.74,.85} & \textbf{349.164 \textbar\ 4.062} \\
\bottomrule
\end{tabular}
\end{table}

We compare five baselines. 
\textbf{B1}~\emph{Raw LLM}: the policy agent with policies $\Pi$ supplied in its prompt and no skill bank $\mathcal{S}$. 
\textbf{B2}~\emph{LLM skills}: $\Pi$ converted into $\mathcal{S}$ by an LLM.
\textbf{B3}~\emph{Human skills}: the same converted $\mathcal{S}$ with domain experts reviewed and edited. 
\textbf{B4}~\emph{Batch update}: $\mathcal{F}$ processed together in a single evolve step.
\textbf{B5}~\emph{AutoSkill}~\citep{autoskill}, a prior skill-evolution method.
B3--B5 and ESCROW all start from the same B2 seed $\mathcal{S}$, $\mathcal{F}$, and base policy agent,
differing only in how $\mathcal{S}$ is improved---human editing (B3),
or skill evolution (B4, B5, ESCROW).
We do not include a baseline that processes feedback one item at a time, since
its $O(|\mathcal{F}|)$ re-evaluation passes are infeasible at production scale.

\begin{figure}
\centering
\includegraphics[width=0.8\linewidth]{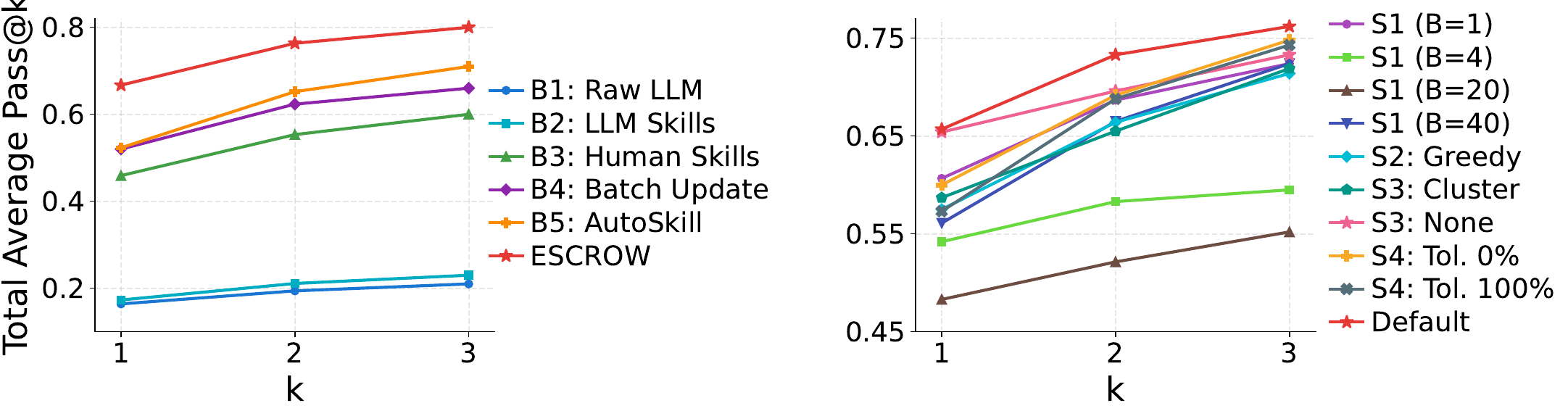}
\caption{Average total pass@$k$ over all cases ($k{=}1$ to $3$) on the held-out FinDAS test set.
\textbf{Left}: baselines and ESCROW (Table~\ref{tab:main}).
\textbf{Right}: ablation settings (Table~\ref{tab:ablation-default}).
ESCROW with its default setting stay on top at every $k$.}
\label{fig:pass-k}
\end{figure}

Table~\ref{tab:main} shows B1 and B2 collapse on hallucination and special cases (to $\le0.043$).
The policy agent fails because the source policies, written for human auditors,
neither abstract rules into a general, non-conflicting form nor make edge cases and grounding explicit.
B3's manual edits recover much of hallucination and special, but relying on intuition
rather than systematic re-scoring, leave coverage uneven and unscalable, at the highest
input cost of any method.
The feedback-driven methods (B4, B5) evolve B2 further, improving monotonically,
with the largest gains on hallucination.
ESCROW reaches the highest overall pass rate, leading the strongest baseline B5
in every category by $0.071$--$0.114$ (paired McNemar $p{=}0.0026$; \S\ref{app:significance}).
Figure~\ref{fig:pass-k} (left) shows the same ordering across $k{=}1{\to}3$, not just
at the reported $k{=}3$. Input cost stays in the same band across methods, while ESCROW has the
lowest output cost; B5 spends $60\%$ more. Since output tokens dominate per-call latency,
ESCROW holds an advantage on both accuracy and deployment latency.
The same pattern transfers to the public $\tau$-bench retail and airline domains, where ESCROW attains the highest or tied-highest overall pass@$k$ at no cost penalty, with its clearest margin on hallucination (airline $1.00$ vs.\ $0.86$; \S\ref{app:taubench}).

\subsection{Ablation study}
\label{sec:ablation}

To quantify the contribution of each module and setting, we sweep four hyperparameters
(S1--S4) one at a time from the default configuration, re-sampling only the evolution set
to 40 cases (same category ratio) to keep the sweep affordable and evolve costs
comparable across S1.
\textbf{S1}~\emph{Batch size} $|\mathcal{B}_i|$: from $1$ (finest) to $40$ (coarsest, whole set in one batch).
\textbf{S2}~\emph{Selection strategy}: the default Pareto frontier ($|\text{frontier}|{\le}3$) versus greedy single-incumbent search ($|\text{frontier}|{\le}1$).
\textbf{S3}~\emph{Consensus filter}: \texttt{section} (default), \texttt{cluster} (stricter), \texttt{none} (no filter).
\textbf{S4}~\emph{Regression tolerance} $\epsilon$: $0\%$ (strict), $5\%$ (default), $100\%$ (unconstrained).

Table~\ref{tab:ablation-default} shows no variant dominates the default: every
alternative lowers total pass@$k$, and the two that beat it in a single category
give up more elsewhere.
Figure~\ref{fig:pass-k} (right) shows the default stays ahead across $k{=}1{\to}3$. 
Batch size \textbf{(S1)} is the most sensitive knob, yet no alternative improves total pass@$k$; 
the finest batching multiplies evolve cost $4.7\times$ in input tokens, 
and the coarsest, cheapest to evolve, carries the
highest inference input cost.
Greedy selection \textbf{(S2)} and both directions of the consensus filter \textbf{(S3)} ---stricter (cluster) and off (none)---all lower total pass@$k$, so frontier
search contributes and \texttt{section} sits at the better operating point; cluster
buys hallucination at a larger loss on general and special.
Neither end of tolerance \textbf{(S4)} improves on the default: $0\%$ rejects useful intermediate edits, while $100\%$
lets any category drop without limit. Selection sees only total accuracy and cost,
so a badly regressed child can still enter the frontier, steering later iterations
into that branch and leaving the frontier populated by points the guard would have rejected.
Inference cost per case otherwise varies little, so the default's lead does not come
from spending more at inference, and the remaining spread falls on the one-time
evolve cost.

\begin{table}[t]
\centering
\caption{Hyperparameter ablations on the FinDAS test set (pass@$k$, $k{=}3$), each block varying one
knob from ESCROW \textbf{Default} settings (top row). Quality cells show the change ($\Delta$ vs.\ default)---\textcolor{green!55!black}{green} gain,
\textcolor{red}{red} drop; cost columns report input\,\textbar\,output tokens in thousands ($k$).}
\label{tab:ablation-default}
\footnotesize
\renewcommand{\arraystretch}{0.95}
\setlength{\tabcolsep}{2.1pt}
\begin{tabular}{@{}c l l cccc cc@{}}
\toprule
& & & \multicolumn{4}{c}{Quality (average pass@$k$, $\Delta$ vs.\ default)} & \multicolumn{2}{c}{Cost (k tokens)} \\
\cmidrule(lr){4-7}\cmidrule(lr){8-9}
\multicolumn{2}{@{}c}{} & Setting & General$\uparrow$ & Hallucination$\uparrow$ & Special$\uparrow$ & Total$\uparrow$
& Inf./case$\downarrow$ & Evol./iter$\downarrow$ \\
\midrule
\multicolumn{3}{@{}l}{\textbf{Default}} & \textbf{0.857} & \textbf{0.829} & \textbf{0.600} & \textbf{0.762}\,\ci{.70,.82} & \textbf{354.632 \textbar\ 4.063} & \textbf{80.213 \textbar\ 24.183} \\
\midrule
\multirow{4}{*}{S1} & \multirow{4}{*}{\shortstack[l]{Batch\\Size}}
 & 1       & 0.829\,\dloss{.028} & 0.829\,\dzero & 0.514\,\dloss{.086} & 0.724\,\dloss{.038}\,\ci{.66,.78} & 402.347 \textbar\ 4.252 & 379.264 \textbar\ 63.418 \\
 & & 4       & 0.571\,\dloss{.286} & 0.757\,\dloss{.072} & 0.457\,\dloss{.143} & 0.595\,\dloss{.167}\,\ci{.53,.66} & 349.285 \textbar\ 4.524 & 113.847 \textbar\ 29.847 \\
 & & 20      & 0.657\,\dloss{.200} & 0.543\,\dloss{.286} & 0.457\,\dloss{.143} & 0.552\,\dloss{.210}\,\ci{.49,.62} & 365.423 \textbar\ 3.942 & 48.315 \textbar\ 15.312 \\
 & & 40      & 0.814\,\dloss{.043} & 0.786\,\dloss{.043} & 0.571\,\dloss{.029} & 0.724\,\dloss{.038}\,\ci{.66,.78} & 453.216 \textbar\ 4.418 & 36.124 \textbar\ 10.847 \\
\midrule
\multirow{2}{*}{S2} & \multirow{2}{*}{\shortstack[l]{Selection\\Strategy}} & \multirow{2}{*}{Greedy} & \multirow{2}{*}{0.857\,\dzero} & \multirow{2}{*}{0.771\,\dloss{.058}} & \multirow{2}{*}{0.514\,\dloss{.086}} & \multirow{2}{*}{0.714\,\dloss{.048}\,\ci{.65,.77}} & \multirow{2}{*}{342.817 \textbar\ 4.214} & \multirow{2}{*}{79.437 \textbar\ 22.418} \\
 & & & & & & & & \\
\midrule
\multirow{2}{*}{S3} & \multirow{2}{*}{\shortstack[l]{Consensus\\Filter}}
 & Cluster & 0.771\,\dloss{.086} & 0.843\,\dgain{.014} & 0.543\,\dloss{.057} & 0.719\,\dloss{.043}\,\ci{.66,.78} & 361.847 \textbar\ 4.038 & 77.834 \textbar\ 22.637 \\
 & & None    & 0.829\,\dloss{.028} & 0.771\,\dloss{.058} & 0.600\,\dzero & 0.733\,\dloss{.029}\,\ci{.67,.79} & 360.124 \textbar\ 3.904 & 82.541 \textbar\ 24.712 \\
\midrule
\multirow{2}{*}{S4} & \multirow{2}{*}{\shortstack[l]{Regression\\Tolerance}}
 & 0\%     & 0.800\,\dloss{.057} & 0.829\,\dzero & 0.614\,\dgain{.014} & 0.748\,\dloss{.014}\,\ci{.69,.80} & 372.918 \textbar\ 4.298 & 79.184 \textbar\ 21.234 \\
 & & 100\%   & 0.829\,\dloss{.028} & 0.800\,\dloss{.029} & 0.600\,\dzero & 0.743\,\dloss{.019}\,\ci{.66,.78} & 382.564 \textbar\ 3.978 & 78.912 \textbar\ 23.574 \\
\bottomrule
\end{tabular}
\end{table}

\begin{figure}
\centering
\includegraphics[width=1\linewidth]{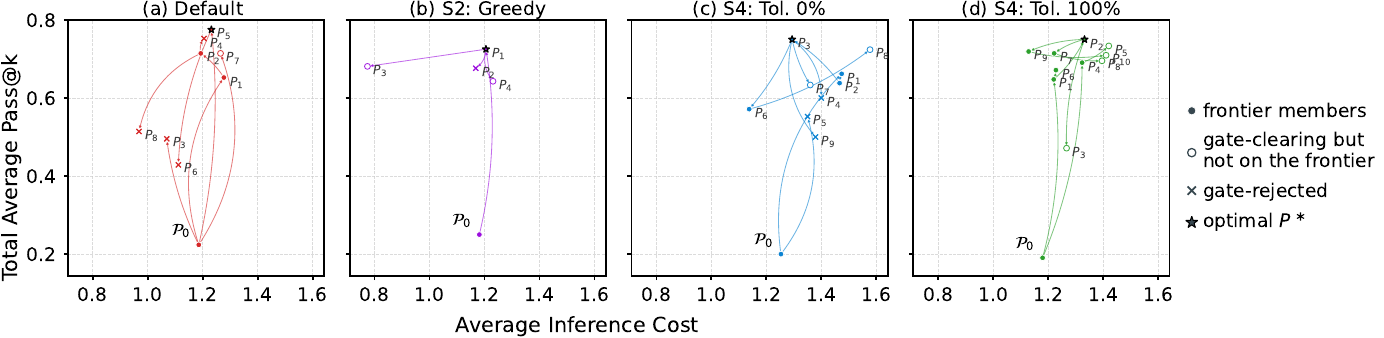}
\definecolor{s4strict}{HTML}{0C83D2}
\definecolor{s4loose}{HTML}{2CA02C}
\definecolor{cfgdefault}{HTML}{D62728}
\definecolor{s2greedy}{HTML}{9C0DDE}

\caption{Candidate admission under default and three ablation settings in \S\ref{sec:ablation}. Cost is inference cost per case in USD, priced from input and output tokens. Filled circles: frontier members; unfilled circles: gate-clearing but not on the frontier; crosses: gate-rejected; stars: optimal $P^\ast$. Arrows connect parents to children. Colour encodes configuration: 
\textcolor{cfgdefault}{default},
\textcolor{s2greedy}{greedy}.
\textcolor{s4strict}{$\epsilon{=}0\%$}, and
\textcolor{s4loose}{$\epsilon{=}100\%$}.
}
\label{fig:admission}
\end{figure}

\subsection{Guarded admission and operating points}\label{sec:admission}
Figure~\ref{fig:admission} traces every candidate produced under the default and three ablation configurations.
Under (a) default, $4$ of $9$ candidates are gate-rejected: each dominates an incumbent on accuracy and cost (Eq.~\ref{eq:dominance}), but its high total pass@$k$ masks a per-category drop below the floor in Eq.~\ref{eq:gate}, buying a gain in one category with another's drop. The closest reaches $0.752$ at cost $\$1.20$ against $\mathcal{P}^*$ at $0.762$.
Rejection keeps a regressed set from seeding later candidates or from being deployed.
(b)  Greedy selection keeps a single incumbent, so mutually non-dominating candidates cannot coexist: where the default evolves from both, greedy retains just the higher pass@$k$ one and discards the cheaper one. Exploration narrows to a single lineage and commits early to $\mathcal{P}^* = 0.714$, exposing a single take-it-or-leave-it point instead of a cost-accuracy menu of operating points.
(c) $\epsilon = 0\%$ leaves no noise headroom and rejects fewer than the default ($3$ of $10$): $\epsilon$ changes which parents the loop evolves from, not just the floor they must clear. Six iterations after $\mathcal{P}^*$ fail to beat it, and the run terminates at $0.748$, below the default---the extra strictness lowers the final evolution quality without a matching safety gain.
(d) $\epsilon = 100\%$ turns admission unconstrained; more candidates qualify as parents to evolve from over the run ($7$ versus $4$), including ones that regressed on a category, yet $\mathcal{P}^*$ stalls at $0.743$---the looser gate does not evolve to a better optimum.

\subsection{Repeated maintenance across update cycles}\label{sec:regression}
\begin{figure}
\centering
\includegraphics[width=0.52\linewidth]{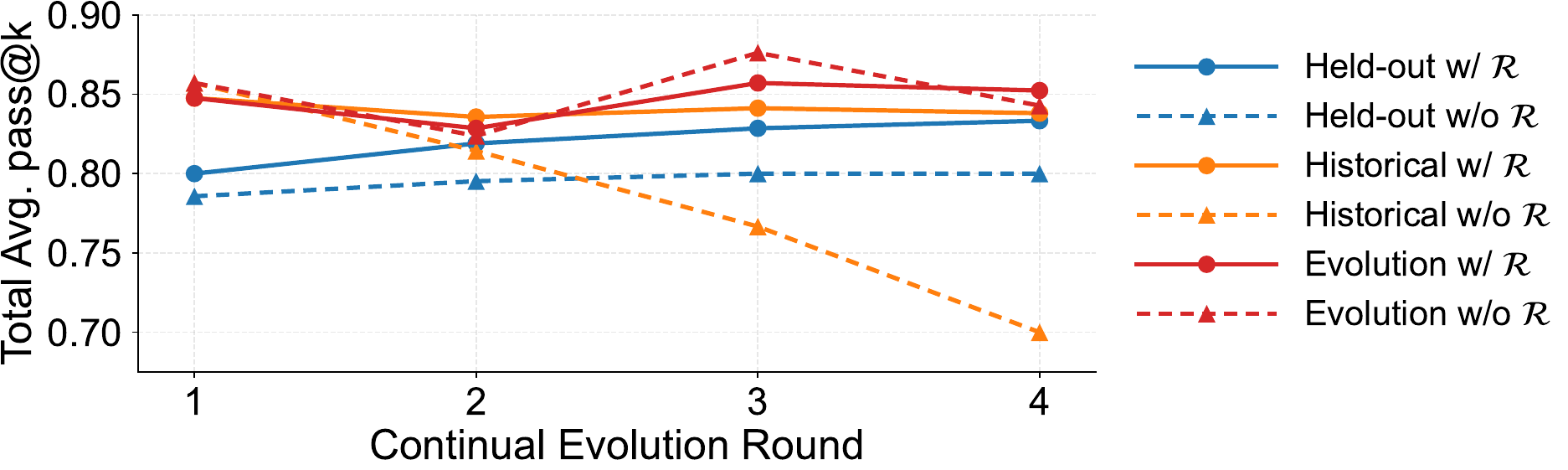}
\caption{Average pass@$k$ over four consecutive evolution rounds on FinDAS,
evaluated on three scopes:
\textcolor{red}{evolution} (sampled and replayed from the window preceding round $t$ that drive that round's evolution),
\textcolor{orange}{historical} (all cumulative evolution cases through round $t$),
and \textcolor{blue}{held-out} ($210$ fixed unseen cases).
Solid lines: full ESCROW (with regression pool $\mathcal{R}$); dashed lines: ESCROW without $\mathcal{R}$.}
\label{fig:regression}
\end{figure}

This section asks not whether maintenance keeps raising accuracy, but whether repeated updates erode what earlier rounds already fixed.
We compare full ESCROW against ESCROW without persistent regression memory ($\mathcal{R}$) over four consecutive rounds, run as a chronological offline replay: four successive $15$-day windows of FinDAS production feedback each supply $210$ cases, replayed identically to both conditions so that every comparison is paired.
Category composition within a window is what production produced rather than stratified, so we report paired totals (\S\ref{app:regression-ci}).
Both conditions start from the same already-evolved bank and thereafter carry forward their own; the only manipulated variable is whether cases resolved in earlier rounds persist into later ones, both as rewrite anchors and in candidate evaluation.
On historical cases (Figure~\ref{fig:regression}), pass@$k$ without $\mathcal{R}$ falls from $0.857$ to $0.7$ over four rounds, showing that later revisions can reintroduce previously resolved failures---the catastrophic forgetting across update cycles that $\mathcal{R}$ is designed to prevent (\S\ref{sec:gate}).
With $\mathcal{R}$, it holds near $0.84$, indicating that persistent historical constraints help retain previously learned behaviour. At the final round the gap is well outside sampling noise (paired difference ${+}0.138$, $95\%$ bootstrap CI $[{+}0.10,{+}0.18]$, excluding zero; \S\ref{app:regression-ci}).
Pass@$k$ on each round's evolution set measures how successfully the system incorporates that round's newly collected cases into its skills. It remains comparable with and without $\mathcal{R}$ across all rounds, indicating that retaining historical cases does not materially limit learning from new ones.
On the held-out set $\mathcal{R}$ scores higher at every round, widening monotonically to ${\approx}0.03$ by $t{=}4$, though the $t{=}4$ paired difference (${+}0.033$ $[{-}0.04,{+}0.10]$) spans zero and is not per-round significant. We read this stable sign as a consistent positive trend---retained history carrying over to unseen cases---rather than a demonstrated held-out gain.

\section{Conclusion}
\label{sec:conclusion}
ESCROW treats a deployed agent's external skills as the unit of \emph{post-deployment
maintenance}, governed by a \emph{Strict Update Boundary}: the LLM proposes revisions, but
only an empirically evaluated version is deployed. Admission is settled by evidence---no
category may fall below its floor, previously resolved cases are retested at every update,
and accuracy is weighed against serving cost---so a change reaches production as a
reviewable diff rather than an unverified rewrite. In production on FinDAS, this yields the
strongest evaluated accuracy--cost trade-off among the methods we compare, and it retains
earlier fixes across repeated update cycles. The dominant bottleneck is per-iteration evaluation cost (see \S\ref{app:limitation} for a full discussion of limitations).
Meanwhile, we acknowledge the potential for algorithmic bias in agent-driven enterprise workflow. 
ESCROW is deliberately designed to reduce this risk and enhance transparency and preserve human accountability:
every behavioural change it makes is a versioned, human-readable natural-language diff that a compliance officer can review and approve.
Nevertheless, any production deployment would still require rigorous, ongoing audits for demographic bias to ensure fair and equitable outcomes.



\clearpage
\phantomsection
\bibliographystyle{abbrvnat}
\bibliography{latex/custom}

@inproceedings{shu2025lava,
  title     = {{LAVA}: Logic-Aware Validation and Augmentation Framework for Large-Scale Financial Document Auditing},
  author    = {Shu, Ruoqi and Wang, Xuhui and Wang, Isaac and Mai, Yanming and Wan, Bo},
  editor    = {Chen, Chung-Chi and Winata, Genta Indra and Rawls, Stephen and Das, Anirban and Chen, Hsin-Hsi and Takamura, Hiroya},
  booktitle = {Proceedings of The 10th Workshop on Financial Technology and Natural Language Processing},
  month     = nov,
  year      = {2025},
  address   = {Suzhou, China},
  publisher = {Association for Computational Linguistics},
  url       = {https://aclanthology.org/2025.finnlp-2.7/},
  doi       = {10.18653/v1/2025.finnlp-2.7},
  pages     = {75--92}
}

@misc{claude-sonnet-4-6,
  title        = {Claude Sonnet 4.6},
  author       = {Anthropic},
  year         = {2026},
  howpublished = {\url{https://www.anthropic.com/news/claude-sonnet-4-6}},
  note         = {Accessed: 2026-06-16}
}

@misc{claude-sonnet-4-5,
  title        = {Claude Sonnet 4.5},
  author       = {Anthropic},
  year         = {2025},
  howpublished = {\url{https://www.anthropic.com}},
  note         = {PLACEHOLDER -- replace with the official model card/reference}
}

@inproceedings{mathew2021docvqa,
  title     = {{DocVQA}: A Dataset for {VQA} on Document Images},
  author    = {Mathew, Minesh and Karatzas, Dimosthenis and Jawahar, C.~V.},
  booktitle = {Proceedings of the IEEE/CVF Winter Conference on Applications of Computer Vision (WACV)},
  pages     = {2200--2209},
  year      = {2021}
}

@article{opsahlong2026officeqa,
  title   = {{OfficeQA} Pro: An Enterprise Benchmark for End-to-End Grounded Reasoning},
  author  = {Opsahl-Ong, Krista and others},
  journal = {arXiv preprint arXiv:2603.08655},
  year    = {2026}
}

@inproceedings{chen2021finqa,
  title     = {{FinQA}: A Dataset of Numerical Reasoning over Financial Data},
  author    = {Chen, Zhiyu and Chen, Wenhu and Smiley, Charese and Shah, Sameena and Borova, Iana and Langdon, Dylan and Moussa, Reema and Beane, Matt and Huang, Ting-Hao and Routledge, Bryan and Wang, William Yang},
  editor    = {Moens, Marie-Francine and Huang, Xuanjing and Specia, Lucia and Yih, Scott Wen-tau},
  booktitle = {Proceedings of the 2021 Conference on Empirical Methods in Natural Language Processing},
  month     = nov,
  year      = {2021},
  address   = {Online and Punta Cana, Dominican Republic},
  publisher = {Association for Computational Linguistics},
  url       = {https://aclanthology.org/2021.emnlp-main.300/},
  doi       = {10.18653/v1/2021.emnlp-main.300},
  pages     = {3697--3711}
}

@inproceedings{yao2024tau,
  title     = {{$\tau$-bench}: A Benchmark for Tool-Agent-User Interaction in Real-World Domains},
  author    = {Yao, Shunyu and Shinn, Noah and Razavi, Pedram and Narasimhan, Karthik},
  booktitle = {International Conference on Learning Representations (ICLR)},
  year      = {2025}
}

@article{chen2021codex,
  title   = {Evaluating Large Language Models Trained on Code},
  author  = {Chen, Mark and Tworek, Jerry and Jun, Heewoo and Yuan, Qiming and Pinto, Henrique Ponde de Oliveira and Kaplan, Jared and Edwards, Harri and Burda, Yuri and Joseph, Nicholas and Brockman, Greg and others},
  journal = {arXiv preprint arXiv:2107.03374},
  year    = {2021}
}

@article{skillsbench,
  title   = {{SkillsBench}: Benchmarking How Well Agent Skills Work Across Diverse Tasks},
  author  = {Li, Xiangyi and Chen, Wenbo and Liu, Yimin and Zheng, Shenghan and Chen, Xiaokun and He, Yifeng and Li, Yubo and You, Bingran and Shen, Haotian and Sun, Jiankai and others},
  journal = {arXiv preprint arXiv:2602.12670},
  year    = {2026}
}

@article{evoskill,
  title   = {{EvoSkill}: Automated Skill Discovery for Multi-Agent Systems},
  author  = {Alzubi, Salaheddin and Provenzano, Noah and Bingham, Jaydon and Chen, Weiyuan and Vu, Tu},
  journal = {arXiv preprint arXiv:2603.02766},
  year    = {2026}
}

@article{autoskill,
  title   = {{AutoSkill}: Experience-Driven Lifelong Learning via Skill Self-Evolution},
  author  = {Yang, Yutao and Li, Junsong and Pan, Qianjun and Zhan, Bihao and others},
  journal = {arXiv preprint arXiv:2603.01145},
  year    = {2026}
}

@article{trace2skill,
  title   = {{Trace2Skill}: Distill Trajectory-Local Lessons into Transferable Agent Skills},
  author  = {Ni, Jingwei and Liu, Yihao and Liu, Xinpeng and Sun, Yutao and Zhou, Mengyu and others},
  journal = {arXiv preprint arXiv:2603.25158},
  year    = {2026}
}

@article{skillx,
  title   = {{SkillX}: Automatically Constructing Skill Knowledge Bases for Agents},
  author  = {Wang, Chenxi and Yu, Zhuoyun and Xie, Xin and Yao, Wuguannan and Fang, Runnan and others},
  journal = {arXiv preprint arXiv:2604.04804},
  year    = {2026}
}

@article{mementoskills,
  title   = {{Memento-Skills}: Let Agents Design Agents},
  author  = {{Memento Team}},
  journal = {arXiv preprint arXiv:2603.18743},
  year    = {2026}
}

@article{skillrl,
  title   = {{SkillRL}: Evolving Agents via Recursive Skill-Augmented Reinforcement Learning},
  author  = {Xia, Peng and Chen, Jianwen and Wang, Hanyang and Liu, Jiaqi and others},
  journal = {arXiv preprint arXiv:2602.08234},
  year    = {2026}
}

@article{metaclaw,
  title   = {{MetaClaw}: Just Talk -- An Agent That Meta-Learns and Evolves in the Wild},
  author  = {Xia, Peng and Chen, Jianwen and Yang, Xinyu and Tu, Haoqin and others},
  journal = {arXiv preprint arXiv:2603.17187},
  year    = {2026}
}

@article{skillzero,
  title   = {{Skill0}: In-Context Agentic Reinforcement Learning for Skill Internalization},
  author  = {Lu, Zhengxi and Yao, Zhiyuan and Wu, Jinyang and Han, Chengcheng and Gu, Qi and Cai, Xunliang and Lu, Weiming and Xiao, Jun and Zhuang, Yueting and Shen, Yongliang},
  journal = {arXiv preprint arXiv:2604.02268},
  year    = {2026}
}

@inproceedings{zwerdling2025policy,
  title     = {Towards Enforcing Company Policy Adherence in Agentic Workflows},
  author    = {Zwerdling, Naama and Boaz, David and Rabinovich, Ella and Uziel, Guy and Amid, David and Anaby Tavor, Ateret},
  editor    = {Potdar, Saloni and Rojas-Barahona, Lina and Montella, Sebastien},
  booktitle = {Proceedings of the 2025 Conference on Empirical Methods in Natural Language Processing: Industry Track},
  month     = nov,
  year      = {2025},
  address   = {Suzhou (China)},
  publisher = {Association for Computational Linguistics},
  url       = {https://aclanthology.org/2025.emnlp-industry.41/},
  doi       = {10.18653/v1/2025.emnlp-industry.41},
  pages     = {595--606}
}

@inproceedings{he2025selfimprove,
  title     = {Enabling Self-Improving Agents to Learn at Test Time With Human-In-The-Loop Guidance},
  author    = {He, Yufei and Li, Ruoyu and Chen, Alex and Liu, Yue and Chen, Yulin and Sui, Yuan and Chen, Cheng and Zhu, Yi and Luo, Luca and Yang, Frank and Hooi, Bryan},
  editor    = {Potdar, Saloni and Rojas-Barahona, Lina and Montella, Sebastien},
  booktitle = {Proceedings of the 2025 Conference on Empirical Methods in Natural Language Processing: Industry Track},
  month     = nov,
  year      = {2025},
  address   = {Suzhou (China)},
  publisher = {Association for Computational Linguistics},
  url       = {https://aclanthology.org/2025.emnlp-industry.115/},
  doi       = {10.18653/v1/2025.emnlp-industry.115},
  pages     = {1625--1653}
}

@inproceedings{zhao2025aitl,
  title     = {Agent-in-the-Loop: A Data Flywheel for Continuous Improvement in {LLM}-based Customer Support},
  author    = {Zhao, Cen and Zhang, Tiantian and Su, Hanchen and Zhang, Yufeng and Su, Shaowei and Xu, Mingzhi and Liu, Yu and Han, Wei and Werner, Jeremy and Cheng, Claire Na and Mehdad, Yashar},
  editor    = {Potdar, Saloni and Rojas-Barahona, Lina and Montella, Sebastien},
  booktitle = {Proceedings of the 2025 Conference on Empirical Methods in Natural Language Processing: Industry Track},
  month     = nov,
  year      = {2025},
  address   = {Suzhou (China)},
  publisher = {Association for Computational Linguistics},
  url       = {https://aclanthology.org/2025.emnlp-industry.135/},
  doi       = {10.18653/v1/2025.emnlp-industry.135},
  pages     = {1919--1930}
}

@article{asda,
  title   = {{ASDA}: Automated Skill Distillation and Adaptation for Financial Reasoning},
  author  = {Yim, Tik Yu and Tan, Wenting and Chan, Sum Yee and Lam, Tak-Wah and Yiu, Siu Ming},
  journal = {arXiv preprint arXiv:2603.16112},
  year    = {2026}
}

@misc{deepagents2024,
  title        = {Deep Agents},
  author       = {{LangChain}},
  year         = {2024},
  howpublished = {\url{https://github.com/langchain-ai/deepagents}},
  note         = {Accessed: 2026-06-15},
}

@article{wei2022chain,
  title={Chain-of-thought prompting elicits reasoning in large language models},
  author={Wei, Jason and Wang, Xuezhi and Schuurmans, Dale and Bosma, Maarten and Xia, Fei and Chi, Ed and Le, Quoc V and Zhou, Denny and others},
  journal={Advances in neural information processing systems},
  volume={35},
  pages={24824--24837},
  year={2022}
}

@inproceedings{yao2023react,
  title     = {{ReAct}: Synergizing Reasoning and Acting in Language Models},
  author    = {Yao, Shunyu and Zhao, Jeffrey and Yu, Dian and Du, Nan and Shafran, Izhak and Narasimhan, Karthik and Cao, Yuan},
  booktitle = {The Eleventh International Conference on Learning Representations},
  year      = {2023},
  url       = {https://openreview.net/forum?id=WE_vluYUL-X}
}

@article{wang2023voyager,
  title   = {Voyager: An Open-Ended Embodied Agent with Large Language Models},
  author  = {Wang, Guanzhi and Xie, Yuqi and Jiang, Yunfan and Mandlekar, Ajay and Xiao, Chaowei and Zhu, Yuke and Fan, Linxi and Anandkumar, Anima},
  journal = {Transactions on Machine Learning Research (TMLR)},
  year    = {2024},
  url     = {https://openreview.net/forum?id=ehfRiF0R3a}
}

@misc{agrawal2026gepareflectivepromptevolution,
      title={GEPA: Reflective Prompt Evolution Can Outperform Reinforcement Learning}, 
      author={Lakshya A Agrawal and Shangyin Tan and Dilara Soylu and Noah Ziems and Rishi Khare and Krista Opsahl-Ong and Arnav Singhvi and Herumb Shandilya and Michael J Ryan and Meng Jiang and Christopher Potts and Koushik Sen and Alexandros G. Dimakis and Ion Stoica and Dan Klein and Matei Zaharia and Omar Khattab},
      year={2026},
      eprint={2507.19457},
      archivePrefix={arXiv},
      primaryClass={cs.CL},
      url={https://arxiv.org/abs/2507.19457}, 
}

@misc{moll2026graspgatedregressionawareskill,
      title={GRASP: Gated Regression-Aware Skill Proposer for Self-Improving LLM Agents}, 
      author={Johannes Moll and Jean-Philippe Corbeil and Jiazhen Pan and Martin Hadamitzky and Daniel Rueckert and Lisa Adams and Keno Bressem},
      year={2026},
      eprint={2605.29668},
      archivePrefix={arXiv},
      primaryClass={cs.AI},
      url={https://arxiv.org/abs/2605.29668}, 
}

@misc{xu2026agentskillslargelanguage,
      title={Agent Skills for Large Language Models: Architecture, Acquisition, Security, and the Path Forward}, 
      author={Renjun Xu and Yang Yan},
      year={2026},
      eprint={2602.12430},
      archivePrefix={arXiv},
      primaryClass={cs.MA},
      url={https://arxiv.org/abs/2602.12430}, 
}

@misc{feizi2026relai,
  title        = {Continual Learning for {AI} Agents: How to Actually Build It},
  author       = {Feizi, Soheil},
  year         = {2026},
  howpublished = {RELAI Blog, \url{https://relai.ai/blog/continual-learning-for-ai-agents}},
  note         = {Accessed: 2026-07-30}
}

@inproceedings{wang2026sage,
  title     = {Reinforcement Learning for Self-Improving Agent with Skill Library},
  author    = {Wang, Jiongxiao and Yan, Qiaojing and Wang, Yawei and Tian, Yijun and Mishra, Soumya Smruti and Xu, Zhichao and Gandhi, Megha and Xu, Panpan and Cheong, Lin Lee},
  booktitle = {Proceedings of the 64th Annual Meeting of the Association for Computational Linguistics (ACL)},
  year      = {2026},
  note      = {arXiv:2512.17102}
}

\clearpage
\appendix
\setlist[enumerate]{leftmargin=1.5em, itemsep=-3pt, topsep=0pt}
\setlist[itemize]{
  leftmargin=1em,  
  itemsep=0pt,     
  topsep=-3pt      
}

\tcbset{
  prompt/.style={
    width=\linewidth,
    colframe=Periwinkle,            
    colback=Lavender!10,            
    colbacktitle=Periwinkle,        
    coltitle=white,                 
    boxrule=0.5pt,
    sharp corners=south,
    fonttitle=\bfseries,
    before upper={\scriptsize\setstretch{1.05}\setlength{\parindent}{0pt}\RaggedRight\setlist[itemize]{topsep=3pt,itemsep=1pt}},
    breakable
  }
}

\newcommand{\variable}[1]{\textcolor{blue}{\texttt{\{#1\}}}}

\newcommand{\skillhead}[1]{\par\addvspace{2pt}\noindent\textbf{#1}\hspace{0.4em}\ignorespaces}

\tcbset{
  skill/.style={
    width=\linewidth,
    colframe=BurntOrange,
    colback=Apricot!15,
    colbacktitle=BurntOrange,
    coltitle=white,
    boxrule=0.5pt,
    sharp corners=south,
    fonttitle=\bfseries,
    before upper={\scriptsize\setstretch{1.05}\setlength{\parindent}{0pt}\RaggedRight\setlist[itemize]{topsep=3pt,itemsep=1pt}\let\paragraph\skillhead},
    breakable
  }
}

\tcbset{
  skill-evolved/.style={
    width=\linewidth,
    colframe=OliveGreen,
    colback=OliveGreen!8,
    colbacktitle=OliveGreen,
    coltitle=white,
    boxrule=0.5pt,
    sharp corners=south,
    fonttitle=\bfseries,
    before upper={\scriptsize\setstretch{1.05}\setlength{\parindent}{0pt}\RaggedRight\setlist[itemize]{topsep=3pt,itemsep=1pt}\let\paragraph\skillhead},
    breakable
  }
}

\tcbset{
  pseudocode/.style={
    width=\linewidth,
    colframe=TealBlue,
    colback=TealBlue!5,
    colbacktitle=TealBlue,
    coltitle=white,
    boxrule=0.5pt,
    sharp corners=south,
    fonttitle=\bfseries,
    before upper={\scriptsize\setstretch{1.05}\setlength{\parindent}{0pt}\RaggedRight\setlist[itemize]{topsep=3pt,itemsep=1pt}},
    breakable
  }
}

\tcbset{
  output/.style={
    width=\linewidth,
    colframe=RoyalBlue,
    colback=RoyalBlue!5,
    colbacktitle=RoyalBlue,
    coltitle=white,
    boxrule=0.5pt,
    sharp corners=south,
    fonttitle=\bfseries,
    before upper={\scriptsize\setstretch{1.05}\setlength{\parindent}{0pt}\RaggedRight\setlist[itemize]{topsep=3pt,itemsep=1pt}},
    breakable
  }
}

\tcbset{
  groundtruth/.style={
    width=\linewidth,
    colframe=ForestGreen,
    colback=ForestGreen!5,
    colbacktitle=ForestGreen,
    coltitle=white,
    boxrule=0.5pt,
    sharp corners=south,
    fonttitle=\bfseries,
    before upper={\scriptsize\setstretch{1.05}\setlength{\parindent}{0pt}\RaggedRight\setlist[itemize]{topsep=3pt,itemsep=1pt}},
    breakable
  }
}

\tcbset{
  errorbox/.style={
    width=\linewidth,
    colframe=BrickRed,
    colback=BrickRed!5,
    colbacktitle=BrickRed,
    coltitle=white,
    boxrule=0.5pt,
    sharp corners=south,
    fonttitle=\bfseries,
    before upper={\scriptsize\setstretch{1.05}\setlength{\parindent}{0pt}\RaggedRight\setlist[itemize]{topsep=3pt,itemsep=1pt}},
    breakable
  }
}

\section{Sample FinDAS documents}

\begin{figure}[h!]
\centering
\begin{subfigure}{0.5\textwidth}
\centering
\includegraphics[width=0.9\linewidth]{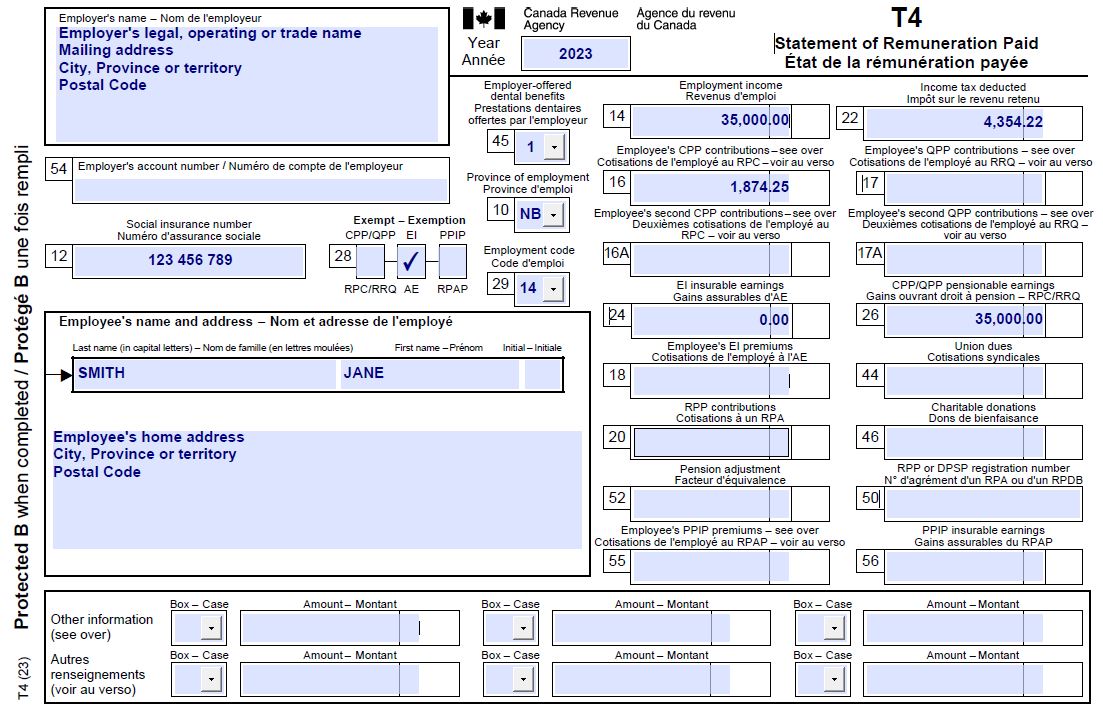}
\caption{Sample tax slip}
\end{subfigure}
\hfill
\begin{subfigure}{0.48\textwidth}
\centering
\includegraphics[width=\linewidth]{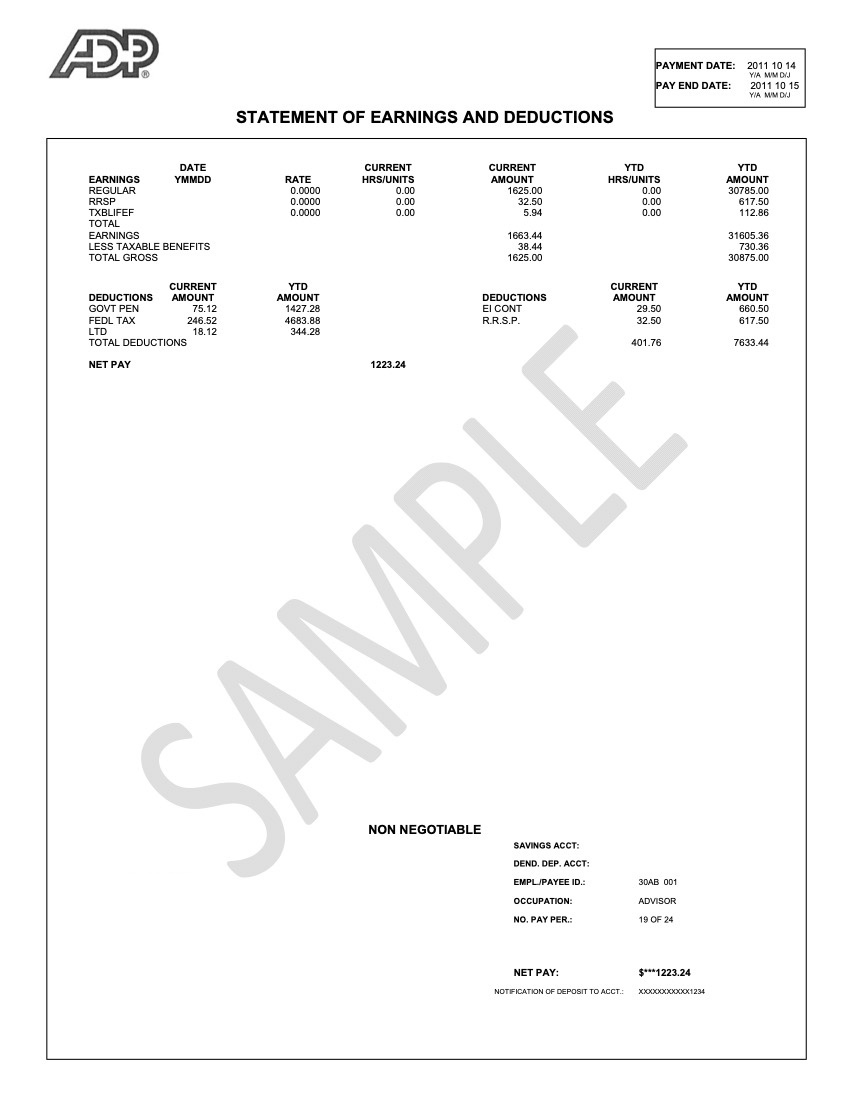}
\caption{Sample paystub}
\end{subfigure}
\caption{Synthetic paystub and tax slip samples used to illustrate the audit skills in \S\ref{app:paystub-skill} and \S\ref{app:t4-skill}.}
\label{fig:fraud-samples}
\end{figure}

Figure~\ref{fig:fraud-samples} shows two synthetic document examples, a paystub and a tax slip, constructed to resemble documents processed by the FinDAS. All contents are fictitious and are included solely for illustrative purposes. Real-world documents exhibit greater diversity in layout, formatting, visual quality, and content complexity.

\FloatBarrier


\section{FinDAS skill corpus and representative examples}
\label{app:skill-corpus}
FinDAS is governed by \textbf{78 policy documents} spanning loan underwriting, payroll auditing, tax compliance, and fraud detection, from which a library of \textbf{over 100 executable skills} is derived across \textbf{23 document types} (e.g., paystubs, tax slips, bank statements, employment letters, and property appraisals). Policy documents vary widely in length and complexity, collectively specifying thousands of compliance requirements at varying levels of granularity --- from single-field validation rules to complex cross-document consistency checks.

To provide concrete insight into skill structure and evolution, we present two representative examples for paystub and tax slip audit. For each, we provide the initial skill derived from policy documentation and the evolved version produced by ESCROW, illustrating how iterative feedback-driven optimization refines grounding rules, edge-case handling, and audit precision while preserving the underlying compliance structure.

\subsection{Paystub auditing skill}
\label{app:paystub-skill}

Below is the initial skill and its evolved successor for paystub auditing. Both follow the same validation structure and output schema; the evolved skill adds more precise grounding rules: an explicit printed-line contract for the \texttt{stated} field, a salary-vs-hourly skip branch, flexible-subset gross-pay reconciliation, and a no-derivation rule for all total fields. The tool groups are sketched in \S\ref{app:tools}.

\begin{tcolorbox}[skill, title={\small\textbf{SKILL.md --- initial}}]
\noindent
\begin{tabular}{@{}>{\bfseries}lp{0.68\linewidth}@{}}
name        & \texttt{paystub-audit-initial} \\[3pt]
description & Verify a paystub for payroll fraud indicators. Triggers on user request and paystub auditing. \\
\end{tabular}

\medskip
\noindent{\bfseries Paystub Audit}

\smallskip
\noindent This skill identifies document inconsistencies, missing fields,
calculation anomalies, and payroll compliance issues.

\medskip
\noindent{\bfseries Inputs}

\begin{itemize}[leftmargin=1.2em,itemsep=0pt,topsep=2pt]
\item \texttt{application\_id}
\item \texttt{document\_type = "paystub"}
\item \texttt{documents} (images + OCR results)
\end{itemize}

\medskip
\noindent{\bfseries\itshape Step 1 \normalfont--- Parse Query and Resolve Paths}

\smallskip
Extract \texttt{application\_id} and \texttt{document\_type}. Call \texttt{`get\_document\_paths(...)`}; stop and report if retrieval fails.

\medskip
\noindent{\bfseries\itshape Step 2 \normalfont--- Prepare Audit Context}

\smallskip
Initialize the fixed-key output skeleton and extract document representations:
\begin{lstlisting}[language=Python,backgroundcolor=\color{Apricot!15},aboveskip=2pt,belowskip=2pt,basicstyle=\ttfamily\scriptsize]
extract_table_structure(image_path)
read_document_image(image_path)
\end{lstlisting}

\medskip
\noindent{\bfseries Check 1 \normalfont--- Field Presence}

\smallskip
A document omitting key fields (employer, payment date, regular pay) may signal manipulation. Call \texttt{`extract\_document\_fields(..., doc\_type="paystub")`}; store each result as \texttt{\{valid, value\}}. \noindent\ldots

\medskip
\noindent{\bfseries Check 2 \normalfont--- Arithmetic Consistency}

\smallskip
Fraudulent edits often update one value without adjusting derived totals. Verify:
\begin{lstlisting}[xleftmargin=0pt,backgroundcolor=\color{Apricot!15},aboveskip=2pt,belowskip=2pt,breaklines=false,basicstyle=\ttfamily\scriptsize]
current_base_pay = rate x hours_worked
current_net_pay  = gross_pay - deductions
\end{lstlisting}
Skip sub-checks where an operand is missing or non-numeric. \noindent\ldots

\medskip
\noindent{\bfseries Check N \normalfont--- Employment Insurance Ceiling}

\smallskip
A YTD EI value above the statutory ceiling is impossible under a legitimate payroll run. Use \texttt{`check\_ei`}. \noindent\ldots

\medskip
\noindent{\bfseries Output}

\smallskip
Fixed-schema JSON, one keyed section per check:
\begin{lstlisting}[xleftmargin=0pt,backgroundcolor=\color{Apricot!15},aboveskip=2pt,belowskip=2pt,basicstyle=\ttfamily\scriptsize]
{
  "expected_fields":  {...},
  "current_base_pay": {...},
  ...,
  "ei": {...}
}
\end{lstlisting}
\end{tcolorbox}
\medskip

\begin{tcolorbox}[skill-evolved, title={\small\textbf{SKILL.md --- evolved}}]
\noindent
\begin{tabular}{@{}>{\bfseries}lp{0.68\linewidth}@{}}
name        & \texttt{paystub-audit-evolved} \\[3pt]
description & Verify a paystub for fraud signals using image and OCR data.
              Triggers on: ``check paystub'', ``verify paystub'', ``paystub fraud'',
              or any paystub fraud-detection request. \\
\end{tabular}

\medskip
\noindent{\bfseries Paystub Auditing}

\smallskip
\noindent Verify a paystub for fraud signals. Work through the checks in order; write each result to the output file before proceeding.

\medskip
\noindent{\bfseries Inputs}

\begin{itemize}[leftmargin=1.2em,itemsep=0pt,topsep=2pt]
\item \texttt{application\_id}
\item \texttt{document\_type = "paystub"}
\item \texttt{documents} (images + OCR results)
\end{itemize}

\medskip
\noindent{\bfseries\itshape Step 1 \normalfont--- Parse Query and Resolve Paths}

\smallskip
Extract \texttt{application\_id} and \texttt{document\_type}. Call \texttt{`get\_document\_paths(...)`}; stop and report if retrieval fails.

\medskip
\noindent{\bfseries\itshape Step 2 \normalfont--- Prepare Audit Context}

\smallskip
Initialize the fixed-key output skeleton and extract document representations:
\begin{lstlisting}[language=Python,backgroundcolor=\color{OliveGreen!8},aboveskip=2pt,belowskip=2pt,basicstyle=\ttfamily\scriptsize]
extract_table_structure(image_path)
read_document_image(image_path)
\end{lstlisting}

\medskip
\noindent{\bfseries Check 1 \normalfont--- Field Presence}

\smallskip
Call \texttt{`extract\_document\_fields(..., doc\_type="paystub")`}, then read image and OCR.

\smallskip
\noindent\textbf{Field extraction rules (evolved):}
\begin{itemize}[leftmargin=1.2em,itemsep=0pt,topsep=2pt]
  \item \texttt{ytd\_regular\_pay}: YTD on the \emph{first/primary earnings line} (``Regular''). Do NOT use the YTD Gross Earnings total.
  \item EI zero vs.\ absent: printed zero $\rightarrow$ \texttt{"0"}; absent $\rightarrow$ \texttt{""}.
\end{itemize}

\noindent\ldots

\medskip
\noindent{\bfseries Check 2 \normalfont--- Arithmetic Consistency}

\smallskip
\noindent\textbf{Sub-check A (evolved):} \texttt{stated} MUST be read from the printed Regular/Base Pay line. Do NOT compute rate$\times$hours for \texttt{stated}; that goes in \texttt{calculation} only.

\smallskip
\noindent\textbf{Salary skip (evolved):} If rate is periodic (``Weekly'', ``Monthly'', ``Salary'') or operands missing: \texttt{calculation: ""}, \texttt{equal: null} --- but still record \texttt{stated} from the printed line.

\smallskip
\noindent\textbf{Sub-checks B \& C (evolved):} Try different subsets of earnings lines before flagging a gross-pay mismatch.

\smallskip
\noindent\textbf{Sub-checks D to G (evolved):} Confirm the labeled total is \textbf{explicitly printed}. If absent: \texttt{calculation: ""}, \texttt{stated: ""}, \texttt{equal: null}. Never infer \texttt{stated}.

\noindent\ldots

\medskip
\noindent{\bfseries Output}

\smallskip
Fixed-schema JSON, one keyed section per check:
\begin{lstlisting}[xleftmargin=0pt,backgroundcolor=\color{OliveGreen!8},aboveskip=2pt,belowskip=2pt,basicstyle=\ttfamily\scriptsize]
{
  "expected_fields":  {...},
  "current_base_pay": {...},
  ...,
  "ei": {...}
}
\end{lstlisting}
\end{tcolorbox}

\subsection{Tax slip audit skill}
\label{app:t4-skill}
Below we show the initial skill and its evolved successor for tax slip auditing. Unlike the paystub skill, whose checks are largely arithmetic (recomputing a stated total from its components), a tax slip follows a fixed layout and its policy is relatively structural: the checks turn on whether specific fields are present or absent rather than on numeric comparison. The evolved skill adds a single-pass efficiency requirement and normalization rules targeting field-extraction failures: monetary value normalization, employer name extraction, standardized address formatting, and a strengthened exemption-conditional box logic grounded solely in the checkbox status.

\begin{tcolorbox}[skill, title={\small\textbf{SKILL.md --- initial}}]
\noindent
\begin{tabular}{@{}>{\bfseries}lp{0.68\linewidth}@{}}
name        & \texttt{tax-slip-audit-initial} \\[3pt]
description & Verify a tax slip for fraud indicators.
              Triggers on ``verify tax slip'', ``check tax slip'',
              or when a tax slip needs fraud review. \\
\end{tabular}

\medskip
\noindent{\bfseries Tax Slip Audit}

\smallskip
\noindent This skill identifies document inconsistencies, missing information, structural anomalies, and compliance issues.

\medskip
\noindent{\bfseries Inputs}

\begin{itemize}[leftmargin=1.2em,itemsep=0pt,topsep=2pt]
\item \texttt{application\_id}
\item \texttt{document\_type = "tax\_slip"}
\item \texttt{documents} (images + OCR results)
\end{itemize}

\medskip
\noindent{\bfseries\itshape Step 1 \normalfont--- Parse Query and Resolve Paths}

\smallskip
Extract \texttt{application\_id} and \texttt{document\_type}. Call \texttt{`get\_document\_paths(...)`}; stop and report if retrieval fails.

\medskip
\noindent{\bfseries\itshape Step 2 \normalfont--- Prepare Audit Context}

\smallskip
Initialize the fixed-key output skeleton and extract document representations:
\begin{lstlisting}[language=Python,backgroundcolor=\color{Apricot!15},aboveskip=2pt,belowskip=2pt,basicstyle=\ttfamily\scriptsize]
extract_document_fields(image_path,doc_type="tax_slip")
read_document_image(image_path)
\end{lstlisting}

\medskip
\noindent{\bfseries Check 1 \normalfont--- Required Fields}

\smallskip
Every legitimate tax slip must contain: tax year, employer identity, employment income, and income tax deducted. Call \texttt{`extract\_document\_fields(...)`}; store each result as \texttt{\{valid, value\}}.

\medskip
\noindent{\bfseries Check 2 \normalfont--- Exemption Consistency}

\smallskip
Tax slips carry three exemption checkboxes. When an exemption is active the corresponding deduction box must be absent or zero. Call \texttt{`extract\_exemption\_status(image\_path)`}; flag any box that carries a value despite the exemption being checked. \noindent\ldots

\medskip
\noindent{\bfseries Check N \normalfont--- Rounding Pattern}

\smallskip
Payroll systems generate values with cents; all-whole-dollar amounts may indicate manual entry. Call \texttt{`check\_rounding(monetary\_values)`}.

\medskip
\noindent{\bfseries Output}

\smallskip
Fixed-schema JSON, one keyed section per check:
\begin{lstlisting}[xleftmargin=0pt,backgroundcolor=\color{Apricot!15},aboveskip=2pt,belowskip=2pt,basicstyle=\ttfamily\scriptsize]
{
  "expected_fields":           {...},
  "exemption_expected_fields": {...},
  ...,
  "box_14":   {...},
  "rounding": {...}
}
\end{lstlisting}
\end{tcolorbox}

\medskip

\begin{tcolorbox}[skill-evolved, title={\small\textbf{SKILL.md --- evolved}}]
\noindent
\begin{tabular}{@{}>{\bfseries}lp{0.68\linewidth}@{}}
name        & \texttt{tax-slip-audit-evolved} \\[3pt]
description & Verify a tax slip for fraud signals using image and OCR data.
              Triggers on: ``check tax slip'', ``verify tax slip'',
              ``tax slip fraud'', or any tax slip fraud-detection request. \\
\end{tabular}

\medskip
\noindent{\bfseries Tax Slip Audit}

\smallskip
\noindent Verify a tax slip for fraud signals and compliance. Work through the checks in order.

\medskip
\noindent{\bfseries Inputs}

\begin{itemize}[leftmargin=1.2em,itemsep=0pt,topsep=2pt]
\item \texttt{application\_id}
\item \texttt{document\_type = "tax\_slip"}
\item \texttt{documents} (images + OCR results)
\end{itemize}

\medskip
\noindent{\bfseries\itshape Step 1 \normalfont--- Parse Query and Resolve Paths}

\smallskip
Extract \texttt{application\_id} and \texttt{document\_type}. Call \texttt{`get\_document\_paths(...)`}; stop and report if retrieval fails.

\medskip
\noindent{\bfseries\itshape Step 2 \normalfont--- Prepare Audit Context}

\smallskip
Initialize the fixed-key output skeleton and extract document representations:
\begin{lstlisting}[language=Python,backgroundcolor=\color{OliveGreen!8},aboveskip=2pt,belowskip=2pt,basicstyle=\ttfamily\scriptsize]
extract_document_fields(image_path,doc_type="tax_slip")
read_document_image(image_path)
\end{lstlisting}

\medskip
\noindent\textbf{Single-pass requirement (evolved):} All checks MUST complete in a single pass. Reuse \texttt{extract\_document\_fields} results across checks; do not re-read the image unnecessarily.

\medskip
\noindent{\bfseries Check 1 \normalfont--- Required Fields}

\smallskip
Use \texttt{`extract\_document\_fields(image\_path, doc\_type="tax\_slip")`}.

\smallskip
\noindent\textbf{Monetary value normalization (evolved):} Apply to every monetary value: (1)~strip thousands-separator commas (``91,999.96'' $\rightarrow$ ``91999.96''); (2)~if OCR returns a whole number but image shows decimal cents, use the image value.

\smallskip
\noindent\textbf{Employer name extraction (evolved):} Record organization name only. Strip from the first street number onward: ``Acme Corp, 123 Main St'' $\rightarrow$ ``Acme Corp''.

\smallskip
\noindent\textbf{Employee address format (evolved):} \texttt{street, city, province, postal-code}. Province as printed; postal code with one internal space; no country code.

\noindent\ldots

\medskip
\noindent{\bfseries Check 2 \normalfont--- Exemption-Conditional Boxes}

\smallskip
Use \texttt{`extract\_exemption\_status(image\_path)`}; reuse Check~1 results for box amounts.

\smallskip
\noindent\textbf{Exemption box logic (evolved):} Validity of each conditional deduction box is determined \emph{solely} by whether the corresponding exemption checkbox is checked. Do not infer exemption from income amount or employment type.

\noindent\ldots

\medskip
\noindent{\bfseries Output}

\smallskip
Fixed-schema JSON, one keyed section per check:
\begin{lstlisting}[xleftmargin=0pt,backgroundcolor=\color{OliveGreen!8},aboveskip=2pt,belowskip=2pt,basicstyle=\ttfamily\scriptsize]
{
  "expected_fields":           {...},
  "exemption_expected_fields": {...},
  ...,
  "box_14":   {...},
  "rounding": {...}
}
\end{lstlisting}
\end{tcolorbox}

\section{Tools for FinDAS agents}
\label{app:tools}

The policy agents call into a fixed FinDAS toolkit (fifteen tools) organized into four functional groups. The groups are described at a high level to convey the toolkit's role concisely, rather than reproducing each tool's full signature and implementation.

\begin{itemize}
\item \textbf{Path resolution and I/O:} locate an application's document files (image, OCR) and persist each checkpoint write to the output file.
\item \textbf{Document understanding:} parse raw document images into structured representations, using tools such as extracting nested table layouts into HTML format for numeric reads, resolving field values into JSON key-value pairs, and detecting discrete binary states (e.g., checkbox fields) via a fine-tuned vision model.
\item \textbf{Arithmetic checks:} recompute a stated total from its components and report whether the two are consistent, skipping gracefully when an operand is missing or non-numeric.
\item \textbf{Statutory and structural checks:} rule-based functions enforce hard compliance constraints, including deduction ceiling checks, field-value digit limits, and pattern-based anomaly detection, producing deterministic valid/invalid verdicts. By delegating such decisions to specialized tools instead of the LLM, the system reduces the likelihood of hallucinated compliance judgments and ensures that decisions are grounded in explicit, auditable validation criteria.
\end{itemize}

\section{ESCROW module prompts}
\label{app:prompts}

Below we present the system prompt and user-message template for each ESCROW module, covering both the cold start (\S\ref{sec:cold_start}) and the evolution loop (\S\ref{sec:retrieval}--\S\ref{sec:propose}); placeholders are written as \variable{variable}.

\subsection{Policy-to-skill splitter and rewriter}
\label{app:prompt-splitter}
The first cold-start module (\S\ref{sec:cold_start}, ``Skill generation''): it consumes the policy corpus $\Pi$ and produces the initial skill bank $\mathcal{S}=\{s_1,\dots,s_L\}$. It runs as two passes. \textbf{Pass~1} (Segmenter) splits each policy document along procedural boundaries---one \emph{verifiable business activity} per segment, defined as one complete audit workflow for one document type (paystub, tax slip, bank statement); individual checks within that workflow are ordered steps inside the same segment, not separate segments. \textbf{Pass~2} (Skill Rewriter) rewrites each segment into a self-contained SKILL.md skill $s_l$ with a description $d_l$ and a back-link to its authoritative clauses in $\Pi$. The two-pass split keeps each LLM call within one audit workflow's scope, which is what makes the resulting SKILL.md self-contained and the segment boundaries auditable.

\begin{tcolorbox}[prompt, title={\small\textbf{System prompt --- Segmenter (Pass 1)}}]
You are a policy segmenter agent. You receive ONE policy document and split it along procedural boundaries into segments, where each segment corresponds to one self-contained unit of verifiable work.

\textbf{What a ``verifiable business activity'' (one segment / one skill) is.} Define the segment boundary by the natural unit of verifiable work in your domain --- typically the artifact / entity / document type that a single end-to-end procedure operates on. Each self-contained procedure corresponds to one such unit. A segment is the COMPLETE procedure for one unit: it bundles every individual step or check that the procedure performs into one ordered whole. Do NOT treat an individual step as its own segment, and do NOT merge two distinct procedures into one segment. Decision rule: if the material is handled as one unit of verifiable work, it is exactly one segment.

\textbf{Boundary rules.}
\begin{itemize}
\item Never split one procedure across two segments; never merge two distinct procedures into one.
\item A segment may span multiple sections of the document; record every section id it covers (across all of its steps).
\item List the individual steps the procedure contains inside the segment summary so the rewrite pass knows which ordered steps to author.
\item Preserve the document's own ordering; segments are emitted in reading order.
\end{itemize}

\textbf{Output schema (JSON only).}
\begin{itemize}
\item \texttt{doc\_id}: the id of the input document.
\item \texttt{segments}: ordered array; each item is \{ \texttt{segment\_id}: ``\variable{doc\_id}-\variable{n}'' (\variable{n} starts at 1, reading order), \texttt{activity}: one short noun phrase naming the single procedure, \texttt{included\_section\_ids}: non-empty array of section identifiers from THIS document that the segment covers (used as the policy back-link), \texttt{summary}: 1--2 sentences describing what the procedure does and its outcome \}.
\end{itemize}

\textbf{Output rules.} JSON only, no prose outside the JSON. \texttt{segments} must be non-empty.
\end{tcolorbox}

\begin{tcolorbox}[prompt, title={\small\textbf{User message --- Segmenter (Pass 1)}}]
\textbf{Policy document} \\
doc\_id: \variable{doc\_id} \\
\variable{policy\_document\_text} \\

\textbf{Task} \\
Split this document into procedural-boundary segments per your instructions. Return JSON only.
\end{tcolorbox}

\begin{tcolorbox}[prompt, title={\small\textbf{System prompt --- Skill Rewriter (Pass 2)}}]
You are a skill author. You receive ONE policy segment and the catalog of tools the procedure may call. You rewrite the segment into a self-contained, executable SKILL.md procedure. The whole workflow is ONE skill.

\textbf{Author the SKILL.md \texttt{content} body with ALL of these sections, in order.}
\begin{itemize}
\item \texttt{name}: name the skill.
\item \texttt{description}: one sentence stating what the skill audits (inconsistencies, missing fields, calculation anomalies, compliance).
\item \texttt{Inputs}: the fields the procedure consumes, one per line (default is \texttt{application\_id}, \texttt{document\_type = ``<type>''}, \texttt{documents} (images + OCR results); if the skill needs more inputs to operate, add them here and prepare them in the setup steps below).
\item \texttt{Setup steps} (Step 1 --- Parse Query and Resolve Paths; Step 2 --- Prepare Audit Context; and Step N for any additional input setup): resolve the document paths, resolve any extra inputs the skill needs to operate, and load the document representations the checks read from, before any check runs; stop and report if path resolution fails.
\item One \texttt{Check <n> --- <name>} per check, in reading order. Each check states (a) the fraud/consistency rule it enforces, grounded in a specific included section; (b) the tool call(s) it makes, by exact name; (c) how it records its result (e.g.\ as \{\texttt{valid}, \texttt{value}\}); and (d) any skip condition (e.g.\ a missing or non-numeric operand).
\item \texttt{Output}: a fixed-schema JSON block with exactly one keyed section per check, in the same order as the checks.
\end{itemize}

\textbf{Grounding and auditability requirements.}
\begin{itemize}
\item Every check must trace to a specific section in the loaded segment source text. Do not invent rules that are not in the segment.
\item Values, choices, and claims the procedure emits must be grounded in the document, a tool result, or a deterministically-derived source --- never fabricated.
\item Record the authoritative section ids as the skill's policy back-link so any agent action is traceable to source policy.
\end{itemize}

\textbf{Tool usage.} Call ONLY tools from the tool catalog provided in the user message; reference each by its exact name and never invent a tool, path, or signature.

\textbf{Output schema (JSON only).}
\begin{itemize}
\item \texttt{name}: \texttt{skill\_id}.
\item \texttt{description}: the frontmatter trigger line (``Verify a paystub for \ldots. Triggers on \ldots''). This is the ONLY field retrieval scores against, so write concrete trigger phrases, not prose.
\item \texttt{content}: the full SKILL.md markdown body with all required sections in order (name, description, Inputs, setup steps, ordered checks, Output).
\end{itemize}

\textbf{Output rules.} JSON only. Keep the skill self-contained and runnable. Do not add rules the segment does not contain, and do not call tools absent from the catalog.
\end{tcolorbox}

\begin{tcolorbox}[prompt, title={\small\textbf{User message --- Skill Rewriter (Pass 2)}}]
\textbf{Policy segment (rewrite this into one skill)} \\
segment\_id: \variable{segment\_id} \\
activity: \variable{activity} \\
summary: \variable{summary} \\
loaded source text (the full text of every included section): \\
\variable{segment\_source\_text} \\

\textbf{Available tools (call ONLY these; reference each by its exact name)} \\
\variable{available\_tools} \\

\textbf{Task} \\
Rewrite this segment into one self-contained SKILL.md procedure per your instructions. Return JSON only.
\end{tcolorbox}

\subsection{Baseline coverage-gap surfacer}
\label{app:prompt-surfacer}
The second cold-start module (\S\ref{sec:cold_start}, ``Feedback triage and bootstrap''). It compares the policy corpus against the current skill bank and emits one base feedback record per uncovered clause; these records are then refined into structured cases by the converter in \S\ref{app:prompt-converter}.

\begin{tcolorbox}[prompt, title={\small\textbf{System prompt --- Baseline Coverage-Gap Surfacer}}]
You are a baseline coverage-gap surfacer of a skill-evolution system. You receive the enterprise policy corpus (as clauses) and the current skill bank (each skill's id, description, and body). Your job is to find policy clauses whose expected behavior is NOT yet exercised by any skill, and emit ONE feedback record per uncovered clause.

\textbf{What to emit per uncovered clause.}
\begin{itemize}
\item \texttt{clause}: the uncovered clause cited from the policy corpus.
\item \texttt{feedback}: a natural-language description of the expected behavior the clause requires and why the current skill bank leaves it unexercised. This is the raw diagnostic seed the converter will expand into the diagnostic record. Do NOT write a rubric, scenario, task, or expected\_output --- those are the converter's job.
\end{itemize}

\textbf{Grounding requirement.} The feedback text must trace to the clause: describe only behavior the clause states. Do not invent requirements the clause does not express.

\textbf{Output schema (JSON only).} \texttt{feedback\_records}: array; each item is \{\texttt{clause}: string, \texttt{feedback}: string\}.

\textbf{Output rules.} JSON only, no prose outside the JSON. One base feedback record per uncovered clause. If no uncovered clause exists, emit \texttt{feedback\_records: []}.
\end{tcolorbox}

\begin{tcolorbox}[prompt, title={\small\textbf{User message --- Baseline Coverage-Gap Surfacer}}]
\textbf{Policy corpus (clauses)} \\
\variable{policy} \\

\textbf{Current skill bank} \\
\variable{skill\_bank} \\

\textbf{Task} \\
Identify clauses whose expected behavior is not yet exercised by the skill bank, and emit one base feedback record per uncovered clause per your instructions. Return JSON only.
\end{tcolorbox}

\subsection{Feedback executability classifier and experiential knowledge}
\label{app:prompt-classifier}

\paragraph{Classifier prompt.}
The executability classifier is a one-time cold-start module (\S\ref{sec:cold_start}) that splits a raw feedback backlog into executable items (forwarded to the converter in \S\ref{app:prompt-converter}) and experiential items (appended to $\mathcal{K}$).

\begin{tcolorbox}[prompt, title={\small\textbf{System prompt --- Feedback Executability Classifier}}]
You are a feedback executability classifier. You receive a backlog of raw operational feedback items and the current skill bank (each skill's id and description). Your job is to label each item by EXECUTABILITY.

\textbf{The executability test.}
Ask: could a fresh agent be handed a self-contained scenario derived from this item, run it end-to-end, and be scored pass/fail against a deterministic rubric?
\begin{itemize}
\item YES $\rightarrow$ label \texttt{"executable"}. The item names a concrete, reproducible behavior or failure on a specific case/document/scenario.
\item NO $\rightarrow$ label \texttt{"experiential"}. The item is general advice, a heuristic, a preference, or a lesson with no replayable setup. It cannot be turned into a scored case, but it is useful as supplementary skill context.
\end{itemize}

\textbf{Match experiential item with skill.}
For \texttt{"experiential"} items: pick the single most relevant \texttt{skill\_id} to attach the advice to, by matching the advice against skill descriptions. If nothing is a good match, set \texttt{target\_skill\_id} to \texttt{null}.

\textbf{Output schema (JSON only).}
\begin{itemize}
\item \texttt{items}: array; each item is \{ \texttt{item\_index}: the input item's index, \texttt{label}: \texttt{"experiential"} $|$ \texttt{"executable"}, \texttt{rationale}: one short sentence justifying the label via the executability test, \texttt{target\_skill\_id}: \texttt{skill\_id} to attach $\mathcal{K}$ to (when label $=$ \texttt{"experiential"}), else \texttt{null} \}.
\end{itemize}

\textbf{Output rules.} JSON only, no prose outside the JSON. Every input item must appear exactly once.
\end{tcolorbox}

\begin{tcolorbox}[prompt, title={\small\textbf{User message --- Feedback Executability Classifier}}]
\textbf{Skill bank (id + description)} \\
\variable{skill\_descriptions} \\

\textbf{Raw operational feedback backlog} \\
\variable{numbered\_raw\_feedback\_items} \\

\textbf{Task} \\
Label each feedback item by executability and map each \texttt{"experiential"} item to its most relevant skill. Return JSON only.
\end{tcolorbox}

\paragraph{Experiential knowledge lifecycle.}
\label{app:experiential-knowledge}
Items classified as \texttt{"experiential"} are stored in a dedicated knowledge library keyed by \texttt{skill\_id}. When a skill is loaded at inference time, its corresponding $\mathcal{K}$ entries are injected into the agent's context alongside the skill body; if the skill is not loaded, its $\mathcal{K}$ is absent from context. $\mathcal{K}$ is append-only and is never subject to the automatic evolution loop, which treats it as read-only context---this prevents unvalidated content from being mutated by the loop or from silently altering scored behavior.

In its current form, $\mathcal{K}$ constitutes a small fraction of feedback (${\sim}3\%$) and is mostly side guidance, so items are simply appended without further processing. As a skill's $\mathcal{K}$ grows, a conflict-resolver or summarizer could be added to reconcile contradictory entries, surface the latest guidance, and cap per-skill length.

$\mathcal{K}$ does not affect the reported experimental results. The evolution loop is driven and scored entirely by the structured cases $(m_i, t_i)$; $\mathcal{K}$ carries no rubric and is never scored. Given its small volume and read-only role, it did not materially enter the evaluated skill sets---it is a deployment convenience for preserving human advice that cannot be turned into a test, rather than a contributor to the measured accuracy--cost numbers.

\subsection{Feedback-to-case converter (human review channel)}
\label{app:prompt-converter}
The converter for the \emph{human review} channel (\S\ref{sec:cold_start}, ``Continuous feedback ingestion''), which refines each groundable reviewer artifact into one structured case $f_i = (m_i, t_i)$ of the form illustrated in \S\ref{app:structuredcase}. The converters for the other two channels (automated trajectory analysis and policy updates) follow the same schema, differing only in the input artifacts they parse; we omit their prompts here for space.

\begin{tcolorbox}[prompt, title={\small\textbf{System prompt --- Feedback-to-Case Converter (Human Review)}}]
You are the converter in a skill-evolution system. You receive raw operational-feedback artifacts and refine each groundable artifact into ONE structured evaluation case.

\textbf{How to build the case.}
\begin{itemize}
\item \texttt{diagnostic\_record}: natural-language description of the failure mode (or the newly required behavior).
\item \texttt{evaluation\_task}: an executable scenario the agent can run:
\begin{itemize}
\item \texttt{category}: one of \texttt{general}, \texttt{hallucination}, \texttt{special}. \emph{hallucination} --- an ungrounded / fabricated value; \emph{special} --- a goal the agent escalated, refused, or halted on instead of reaching via an in-policy path, or an edge case not covered by policy; \emph{general} --- a largely-correct case with a minor correction.
\item \texttt{description}: the user-facing task statement, reusing the artifact's \texttt{application\_id} where one exists.
\item \texttt{task}: \{ \texttt{task\_id}, \texttt{evaluation\_criteria} \} --- the rubric must encode the corrected / amended behavior precisely enough to score a fresh agent run pass/fail.
\item \texttt{expected\_output}: the ground truth; if the exact key and value are not provided by the raw feedback, return \texttt{null}.
\end{itemize}
\end{itemize}

\textbf{Grounding requirement.} Every rubric criterion and \texttt{expected\_output} value must trace to the artifact. Do not invent requirements the artifact does not state. If an artifact's ground truth cannot be reconstructed, SKIP it --- do not guess.

\textbf{Output schema (JSON only).}
\begin{itemize}
\item \texttt{cases}: array; each item is \{ \texttt{diagnostic\_record}: string, \texttt{evaluation\_task}: \{ \texttt{category}, \texttt{description}, \texttt{task}: \{ \texttt{task\_id}, \texttt{evaluation\_criteria} \}, \texttt{expected\_output} \} \}.
\item \texttt{skipped}: array of \{ \texttt{feedback\_id}: reason \} for artifacts that are not groundable into a scorable case.
\end{itemize}

\textbf{Output rules.} JSON only, no prose outside the JSON. Emit at most one case per artifact.
\end{tcolorbox}

\begin{tcolorbox}[prompt, title={\small\textbf{User message --- Feedback-to-Case Converter (Human Review)}}]
\textbf{Operational-feedback artifacts} \\
\variable{numbered\_feedback\_artifacts} \\

\textbf{Task} \\
Refine each artifact into a structured case, or skip it if it is not groundable into a scorable rubric. Return JSON only.
\end{tcolorbox}

\subsection{Diagnoser (per feedback batch)}
\begin{tcolorbox}[prompt, title={\small\textbf{System prompt --- Diagnoser}}]
You are one of $N$ parallel diagnoser agents in an evolution loop. You will receive a parent skill program (a set of skills) and a batch of feedback items (failed and/or passed cases executed against that program).

\textbf{Your job.} (1)~Extract generalizable success patterns from passing baseline items (\texttt{general}); these are what the program already gets right and must be preserved. (2)~Analyze the dominant root cause(s) across the failing items (\texttt{special} and \texttt{hallucination}). (3)~Propose one short paragraph of edit plan per skill that should be edited to fix those root causes without breaking the success patterns.

\textbf{Inputs.} \emph{Parent Program:} a list of skills' main content (SKILL.md), each with \texttt{skill\_id}, \texttt{description}, and full markdown body. \emph{Feedback Batch:} a numbered list with \texttt{category}, \texttt{description}, \texttt{evaluation} (per-check outcomes: \texttt{db\_check}, \texttt{action\_checks}, \texttt{communicate\_checks}, and \texttt{static\_json} verdict with field-level mismatches), optional \texttt{incomplete} flag (set when the agent hit \texttt{max\_steps}), and \texttt{annotation} for non-\texttt{general} items.

\textbf{Output schema (enforced by JSON schema).}
\begin{itemize}
\item \texttt{categories}: subset of \{\texttt{special}, \texttt{hallucination}\} addressed by the proposed edit.
\item \texttt{root\_cause}: short paragraph identifying the dominant failure pattern, grounded in failing items and contrasted against \texttt{general} success patterns.
\item \texttt{plans}: object mapping each target \texttt{skill\_id} to one short paragraph describing what sections to add, change, or remove, and why. Keys must be a subset of the parent's \texttt{skill\_id}s.
\end{itemize}

\textbf{Per-category guidance.}
\begin{itemize}
\item \emph{special} --- the agent escalated, refused, or halted (including running out of step budget) before reaching the user's underlying goal through the in-policy paths available. Push the SKILL.md toward completing the user's goal through in-policy paths: restate the goal in policy terms, enumerate every in-policy branch that could advance it, attempt those options in policy-prescribed order, prefer batch tool forms over repeating the single-item form, escalate or refuse only the residual portion that has no in-policy path.
\item \emph{hallucination} --- the agent's actions or outputs diverge from what the user, policy, or available sources actually specified. Push the SKILL.md toward grounding behavior in explicit, traceable sources before acting or replying: every value, choice, and claim should trace to a specific user utterance, policy clause, or deterministically-derived source; user-given values should override policy defaults when policy permits; qualitative or referential expressions should be resolved into concrete values; when grounding is missing, ask or verify rather than fill the gap.
\item \emph{general} --- standard routine cases that are primarily correct, with only minor errors. The errors still warrant repair, but correct ones supply success patterns to preserve: extract what the SKILL.md already gets right and ensure the edits proposed for the other categories do not regress this behavior.
\end{itemize}

\textbf{Targets.} An edit plan can target either the SKILL.md prose (add a fallback/retry loop, verification checkpoint, edge-case rule, workflow step) or a bundled script the skill invokes (fix a bug, add error handling, harden a network call) referenced by its existing path.

\textbf{Output rules.} \texttt{plans} must be non-empty (an empty object is rejected); be concise but specific; do not output reasoning outside the JSON.
\end{tcolorbox}

\begin{tcolorbox}[prompt, title={\small\textbf{User message --- Diagnoser}}]
\textbf{Parent Program} \\
\variable{rendered\_parent\_skills} \\

\textbf{Feedback Batch} \\
\variable{rendered\_feedback\_batch} \\

\textbf{Task} \\
Produce ONE proposal in the schema above. JSON only.
\end{tcolorbox}

\subsection{Consolidator}
\begin{tcolorbox}[prompt, title={\small\textbf{System prompt --- Consolidator}}]
You are a consolidator agent in an evolution loop. You receive $N$ peer-reviewed proposals from parallel diagnoser agents who each analyzed a different feedback batch from the same parent skill program. A consolidation strategy has already filtered the proposals down to a ``surviving'' set you must synthesize over.

\textbf{Your job.} Produce ONE consolidated edit plan per surviving \texttt{skill\_id} that captures the agents' consensus while dropping idiosyncratic / one-off suggestions.

\textbf{Inputs.} \texttt{surviving\_skill\_ids} (the exact set of skills you must produce plans for); per agent: \texttt{root\_cause} and \texttt{plans} entries restricted to surviving skills.

\textbf{Output schema.}
\begin{itemize}
\item \texttt{categories}: non-empty subset of \{\texttt{special}, \texttt{hallucination}, \texttt{general}\}, reflecting the union addressed across the agents.
\item \texttt{root\_cause}: a concise paragraph summarizing the dominant failure pattern the agents agree on; no idiosyncratic claims.
\item \texttt{plans}: object whose keys equal \texttt{surviving\_skill\_ids} (no extras, no omissions); each value is one focused paragraph that merges concrete suggestions from multiple agents (preserving tool names, file paths, conditional logic), drops one-off / contradictory suggestions, and is self-contained enough that a downstream LLM can rewrite the SKILL.md from this paragraph alone.
\end{itemize}

\textbf{Output rules.} JSON only, no prose outside the JSON.
\end{tcolorbox}

\begin{tcolorbox}[prompt, title={\small\textbf{User message --- Consolidator}}]
\textbf{Surviving skill\_ids} \\
\variable{surviving\_skill\_ids\_json\_array} \\

\textbf{Agents' proposals (restricted to surviving skills)} \\
\variable{rendered\_proposals\_filtered} \\

\textbf{Task} \\
Synthesize a single consolidated proposal. JSON only.
\end{tcolorbox}

\subsection{Skill builder (rewriter)}
The skill-rewriting stage (\S\ref{sec:propose}), applied to one target skill at a time. Sibling skills are passed as context-only (frontmatter \texttt{description} only, not full bodies) so the rewriter knows what other skills are responsible for and avoids duplication. Auxiliary files in the skill folder are listed by relative path so the rewriter can reference them without inventing paths.

\begin{tcolorbox}[prompt, title={\small\textbf{System prompt --- Skill Builder}}]
You are a skill rewriter in an evolution loop. Skills are STRUCTURED PROCEDURES, not free text. They typically contain if-then branching for input variants, explicit tool calls (e.g.\ \texttt{table\_to\_html}, \texttt{find\_evidence}, \texttt{verify\_claim}), multi-step pipelines with intermediate checks, and verification / grounding steps before final output.

\textbf{Inputs.} The target skill (frontmatter + full markdown body --- rewrite this); sibling skills in the parent program (context only, do not edit them); a consolidated edit plan describing the change to apply.

\textbf{Auxiliary files.} Aux files (\texttt{scripts/}, \texttt{references/}, \texttt{agents/}, \texttt{assets/}) live alongside SKILL.md on disk and are copied to the new skill folder unchanged. You may reference them in the rewritten content using their existing relative paths; do not invent paths that do not exist in the parent.

\textbf{Versioning.} The skill is being versioned. If the parent's markdown body references the \texttt{parent\_skill\_id} anywhere (H1 title, section headings, self-references), replace those occurrences with \texttt{new\_skill\_id}.

\textbf{How to edit.} Apply the edit plan precisely. Preserve the skill's overall structure (section headings, ordering, existing tool calls) unless the plan explicitly says otherwise. Keep the skill self-contained and runnable.

\textbf{Output.} Return ONE JSON object with: \texttt{description} (the trigger / when-to-use line for the frontmatter) and \texttt{content} (the full updated markdown body). No YAML frontmatter, no outer code fences, no commentary or reasoning text outside the JSON.
\end{tcolorbox}

\begin{tcolorbox}[prompt, title={\small\textbf{User message --- Skill Builder}}]
\textbf{Target skill (rewrite this)} \\
\variable{rendered\_target\_skill} \\

\textbf{Versioning} \\
parent\_skill\_id: \variable{parent\_skill\_id} \\
new\_skill\_id: \variable{new\_skill\_id} \\
Replace any occurrence of \variable{parent\_skill\_id} in the body (H1 title, headings, self-references) with \variable{new\_skill\_id}. \\

\textbf{Auxiliary files in target skill folder (do not invent new paths)} \\
\variable{aux\_file\_list} \\

\textbf{Sibling skills (context only)} \\
\variable{sibling\_skill\_descriptions} \\

\textbf{Consolidated edit plan (for \variable{target\_skill\_id})} \\
categories: \variable{categories} \\
root\_cause: \variable{root\_cause} \\
plan: \variable{plan\_for\_target} \\

\textbf{Task} \\
Rewrite the target skill to fully apply the edit plan above. Respond with JSON only.
\end{tcolorbox}

\section{From operation feedback to structured cases}
\subsection{Raw feedback corpus samples}
\label{app:feedback-samples}
These samples are drawn from the operational feedback that FinDAS accumulates as its policy agents execute skills on real cases. As described in \S\ref{sec:cold_start}, feedback enters through three channels --- human review, automated trajectory analysis, and policy-update diffs --- and each is converted into a diagnostic record that states the failure mode in free-form natural language. Below we show representative records from each category.

\paragraph{Special}
\begin{itemize}[itemsep=3pt]
  \item When an exemption box is checked, the corresponding value should be either blank or zero.
  \item When no YTD Net Pay line appeared on the document, the agent attempted to derive the figure from other fields rather than treating the absence as a skip condition --- if the relevant printed line does not exist, the check should be omitted entirely.
\end{itemize}

\paragraph{Hallucination}
\begin{itemize}[itemsep=3pt]
  \item The stated base pay was computed as rate $\times$ hours rather than read from the printed Regular/Base Pay line. Even when rate or hours are absent, the printed line value should still be recorded; only the arithmetic cross-check is skipped.
  
  \item The stated value for total deductions was re-summed from individual line items rather than read from the printed ``Total Deductions'' line. On documents that separate Taxes from Payroll Deductions into distinct subtotals, the relevant figure is the Payroll Deductions subtotal, not the combined amount.
  \item Gross pay was flagged as a mismatch after trying only one combination of earnings lines. Different documents include different subsets of earnings components; all plausible combinations should be explored before flagging a discrepancy.
\end{itemize}

\paragraph{General} 
\begin{itemize}[itemsep=3pt]
\item Monetary values containing thousands-separator commas (e.g., \texttt{"90,169.00"}) were not normalized --- the comma should be stripped before any numerical comparison.
  \item The employee address included a country suffix (e.g., ``CAN'') that should be omitted, and the postal code was missing the required internal space (e.g., \texttt{"M5V~2T6"}, not \texttt{"M5V2T6"}).
\end{itemize}

\subsection{Structured case example}
\label{app:structuredcase}

Below we show a single structured case $f_i = (m_i, t_i)$ converted from human review notes drawn from the FinDAS. Here $m_i$ is the natural-language diagnostic record identifying the failure mode, and $t_i$ is the executable evaluation task.

\begin{tcolorbox}[output, title={\small\textbf{Structured case $f_i$}}]
\textbf{$m_i$} = \textquotedblleft{}The base pay stated value must be read directly from the printed Regular/Base Pay line. The formula rate $\times$ hours serves only as a cross-check and must not substitute the printed value. When rate or hours are absent, the check is skipped but the printed value is still recorded. YTD net pay must likewise come from the printed YTD Net Pay line; if absent, the field is left blank.\textquotedblright

\medskip
\noindent\textbf{$t_i$} = \{

\quad \texttt{"category":} \textquotedblleft{}hallucination\textquotedblright,

\quad \texttt{"description":} \textquotedblleft{}Verify application \variable{ID}'s paystub documents for potential fraud \ldots\textquotedblright,

\quad \texttt{"task":} \{

\qquad \texttt{"task\_id":} 2,

\qquad \texttt{"evaluation\_criteria":} \{ 

\qquad\quad \ldots

\qquad\quad \textit{// criteria by which task success is determined}

\qquad \}

\quad \},

\quad \texttt{"expected\_output":} \{ 

\qquad \ldots

\qquad \textit{// ground-truth output the agent must produce to earn reward}

\quad \}

\}
\end{tcolorbox}

\section{Experiment setting}
\label{app:config}
\paragraph{Operating-point selection.}Because FinDAS operates in a high-stakes financial setting, a wrong verdict carries far greater downstream cost than a few extra tokens; 
we therefore prioritize accuracy ($\mathrm{PR}$) over inference cost when selecting $\mathcal{P}^{*}$ in experiments, breaking ties toward lower cost (Table~\ref{tab:config}).
A live deployment may instead pick any frontier point to match its risk and budget (\S\ref{sec:frontier}).
The baselines carry no such setting. 
B1-B3 perform no evolution, B4 runs a single evolve step with one output, and B5 emits the final skill set of its own mechanism.

Table~\ref{tab:config} reports the default ESCROW configuration used throughout the
experiments, except where an ablation in \S\ref{sec:ablation} varies a single knob.

\newcommand{\reprow}[2]{\quad #1 \dotfill\ #2 \\}
\newcommand{\repgroup}[1]{\multicolumn{1}{@{}l}{\emph{#1}} \\[1pt]}

\begin{table}[t]
\centering
\caption{Default configuration of ESCROW in the \S\ref{sec:experiment}}
\label{tab:config}
\small
\setlength{\tabcolsep}{0pt}
\renewcommand{\arraystretch}{1.25}
\begin{tabular}{@{}p{\linewidth}@{}}
\toprule
\textbf{Parameter} \hfill \textbf{Default value} \\
\midrule
\repgroup{Models}
\reprow{Diagnoser, Consolidator, Skill Builder}{Claude Sonnet~4.6 (thinking enabled)}
\reprow{Cold-start Segmenter, Rewriter, Surfacer, Converter}{Claude Sonnet~4.6 (thinking enabled)}
\reprow{Embedder (cold-start retrieval)}{Titan Embed Text~v2}
\reprow{Policy agent (FinDAS evaluation)}{Claude Sonnet~4.6 (thinking enabled)}
\reprow{Policy agent \& user simulator ($\tau$-bench evaluation)}{Claude Sonnet~4.5 (thinking enabled)}
\addlinespace[2pt]


\addlinespace[2pt]
\repgroup{Evolution loop}
\reprow{Top-$k$}{$5$}
\reprow{Diagnoser batch size $|\mathcal{B}_i|$}{$8$ ($N{=}\lceil|\mathcal{F}|/8\rceil$ parallel diagnosers)}
\reprow{Max iterations $T$}{$10$}
\reprow{Convergence patience $\tau$}{$3$ iterations without improvement}
\reprow{Maximum frontier size $|\textit{frontier}|$}{$\leq 3$ (Pareto non-dominated set)}
\reprow{$\mathcal{P}^{*}$ Selection}{ $\arg\max_{\mathcal{P}\in\textit{frontier}}(\mathrm{PR},-\mathrm{Cost})$ (accuracy-first)}
\reprow{Consolidation strategy}{\texttt{section} (default); options: \texttt{section}, \texttt{cluster}, \texttt{none}}


\reprow{Voting threshold}{$40\%$ of diagnosers}
\reprow{Regression tolerance $\epsilon$}{5\% (relative, per category)}
\reprow{Regression pool sample per iteration}{sized in proportion (50\%) to the task size in the current evolution round}
\reprow{Batch shuffle seed}{per-iter $=$ base $+$ iter (base $=0$)}


\bottomrule
\end{tabular}
\end{table}

\section{Experiment discussion}
\subsection{Data sources and sampling design}
\label{app:dataset}
Evaluating what ESCROW targets requires more than a set of scored tasks:
a governing policy the skills are derived from and traced back to,
multi-step tool-using execution scored on the final verdict,
a record of how a deployed agent actually failed, independent of the method under test,
and enough cases twice over---one set for the loop to evolve from,
and a disjoint held-out set, large enough per risk category, to validate what it produces.
No public resource we are aware of provides all these together,
which is what makes an internal system necessary here rather than preferred.
Document-QA benchmarks supply tasks as document--question pairs but no governed workflow to execute,
and no ground truth over an agent's actions to score that execution against (\S\ref{sec:setup}).
Policy-governed agent benchmarks supply the policy and the workflow---enough for ESCROW to cold-start---
but their cases carry gold answers rather than failure histories,
and a method scored on signal it synthesized for itself
cannot demonstrate that real corrections generalize into reusable rules.
Among the latter, $\tau$-bench is the closest match and the harness we build on (\S\ref{app:taubench});
its domains probe generality rather than substituting for FinDAS,
their pools ($110$ and $50$ cases) being thin once split into an evolution set and a per-category held-out set.
FinDAS supplies the missing combination:
$78$ policy documents across $23$ document types (\S\ref{app:skill-corpus}),
reviewer feedback accumulated from real audits,
and a case pool deep enough to fill three equal categories twice over at the size used below.

The evolution set and the held-out test set are both drawn from the FinDAS case pool
and stratified into equal category cells ($70$ general, $70$ hallucination, $70$ special each) 
rather than sampled at operational incidence, under which hallucination-type risk dominates.
Each case carries the category its converter assigns from the reviewer's annotation (\S\ref{app:prompt-converter}), 
verified by a domain expert.
The balance is a measurement choice: 
the reported unit is the per-category pass rate $\mathrm{PR}_c$ and the gate holds a floor in each category independently (\S\ref{sec:gate}), 
so unequal cells would estimate the categories at different resolutions and let the most frequent one carry the total.
With equal cells one case moves a category rate by $1/70 \approx 0.014$ and the total by ${\approx}0.005$,
so the differences we report are not single-case artifacts.
The cell size stops at $70$ because each case is a full multi-trial agent run and the evolution set is re-scored for every candidate at every iteration, 
making evaluation---not sampling---the binding cost.
The two sets are disjoint and the held-out set takes no part in the loop: 
diagnosis, the non-regression gate, and the frontier selection of $\mathcal{P}^{*}$ all read the evolution set and the regression pool only, 
and the held-out set is scored once per method in experiments.

\subsection{Significance and structure of the improvement}
\label{app:significance}
ESCROW leads the strongest baseline \textbf{B5} in every category
(Table~\ref{tab:main}) and on total at every $k$
(Figure~\ref{fig:pass-k}), and its widest margin falls on hallucination---the
dominant safety risk in this financial setting---where it also attains its
highest absolute pass rate.

In Table~\ref{tab:main} the interval for ESCROW ($[.74,.85]$) overlaps that of the
strongest baseline \textbf{B5} ($[.65,.77]$). These intervals are marginal---each
computed for one method on its own---so their overlap does not imply that the
difference between the two is zero, and we test that difference directly.
The comparison here is paired: both methods are scored on the same $210$ held-out cases, 
and with $n{=}k{=}3$ each case yields a binary pass@$3$, 
so the two runs can be compared case by case. Pairing is what makes the test sharp. 
Of the $210$ cases, $173$ resolve identically under both methods, leaving $37$ discordant cases to carry the signal;
variation in case difficulty moves both methods together and cancels in the difference.
We test $H_0$, that the two methods have equal pass@$3$ on the test distribution, against $H_1$, that they differ. 
The observed gap is $\Delta{=}{+}0.090$ ($0.800$ vs.\ $0.710$, or $19$ additional cases passed), 
and a paired bootstrap over cases ($B{=}10{,}000$) puts its $95\%$ interval at $[{+}0.038,{+}0.148]$, which excludes zero. 
McNemar's exact test reaches the same verdict from the discordant cases alone: 
$28$ of them are won by ESCROW and $9$ by \textbf{B5}, whereas under $H_0$ each should fall either way with equal probability,
and a split this lopsided or more has two-sided probability $p{=}0.0026$. 
We therefore reject $H_0$ at the $5\%$ level: the lead over the strongest baseline is
not attributable to sampling. It is also not an artifact of the reported $k$---the
paired interval excludes zero at $k{=}1$, $2$, and $3$ ($\Delta{=}{+}0.144$,
${+}0.111$, ${+}0.090$)---nor specific to \textbf{B5}: against \textbf{B4}, the next
strongest baseline, the margin is larger and the evidence stronger
($\Delta{=}{+}0.143$, two-sided $p{=}2.8\times10^{-6}$).

That the lead is broad rather than concentrated is
what the per-category gate produces (\S\ref{sec:gate}).
Because \textbf{B5} is itself an evolved skill bank from the same seed
$\mathcal{S}$, feedback $\mathcal{F}$, and policy agent, this margin measures what
ESCROW adds over a prior skill-evolution method rather than whether skill evolution
helps at all; Table~\ref{tab:ablation-default} separates the individual
contributions of frontier search, consensus filtering, and the gate. The gain carries no
cost premium: ESCROW posts the lowest output cost of any method, at an input
cost within the range spanned by the baselines.

\subsection{Uncertainty in the multi-cycle experiment}
\label{app:regression-ci}
We attach the same statistical protocol as the main table to the multi-cycle study
(\S\ref{sec:regression}): $95\%$ percentile bootstrap CIs by case-level resampling
($B{=}10{,}000$). Because both conditions are scored on identical case sets at every
round, all cross-condition comparisons are \emph{paired}: we resample cases, take the
per-case difference of pass@$3$ indicators, and read the $2.5$/$97.5$ percentiles of the difference distribution. 
Evolution-set CIs are computed on $\mathcal{F}_t$, the $210$ FinDAS production cases sampled and replayed from the window preceding round $t$ that drive that round's evolution;
historical-case CIs on $\mathcal{H}_t = \bigcup_{i=1}^{t} \mathcal{F}_i$
($210$, $420$, $630$, $840$ at $t{=}1{\dots}4$); 
held-out CIs on the fixed $210$ cases.
To avoid clutter we omit error bars from Figure~\ref{fig:regression} and report the intervals here instead.
Unlike the stratified evolution and held-out sets (\S\ref{app:dataset}), each replayed window carries whatever category mix production actually yielded, which varies across rounds.
Because both conditions are scored on the same cases, every paired comparison is unaffected; but per-category cells differ in size from round to round, so we report per-round totals and read category-level evidence from Table~\ref{tab:main} and \S\ref{sec:admission}.

\begin{table}[htbp]
\centering
\caption{Per-round total average pass@$3$ across all categories with $95\%$ bootstrap CIs for the multi-cycle experiment
(\S\ref{sec:regression}). Both conditions share the initial skill bank and are scored on
identical case sets each round.}
\label{tab:regression-ci}
\footnotesize
\setlength{\tabcolsep}{4pt}
\begin{tabular}{@{}l l cccc@{}}
\toprule
Scope & Cond. & $t{=}1$ & $t{=}2$ & $t{=}3$ & $t{=}4$ \\
\midrule
\multirow{2}{*}{Held-out}
 & with $\mathcal{R}$      & .800\,[.74,.85] & .819\,[.77,.87] & .829\,[.78,.88] & .833\,[.78,.88] \\
 & without $\mathcal{R}$   & .786\,[.73,.84] & .795\,[.74,.85] & .800\,[.74,.85] & .800\,[.74,.85] \\
\midrule
\multirow{2}{*}{Historical}
 & with $\mathcal{R}$     & .848\,[.80,.90] & .836\,[.80,.87] & .841\,[.81,.87] & .838\,[.81,.86] \\
 & without $\mathcal{R}$  & .857\,[.81,.90] & .814\,[.78,.85] & .767\,[.73,.80] & .700\,[.67,.73] \\
\midrule
\multirow{2}{*}{Evolution}
 & with $\mathcal{R}$     & .848\,[.80,.90] & .829\,[.78,.88] & .857\,[.81,.90] & .852\,[.80,.90] \\
 & without $\mathcal{R}$  & .857\,[.81,.90] & .824\,[.77,.88] & .876\,[.83,.92] & .843\,[.79,.89] \\
\bottomrule
\end{tabular}
\end{table}

On historical cases the two conditions are indistinguishable at $t{=}1$
(overlapping CIs) and separate monotonically thereafter; by $t{=}4$ the intervals are
disjoint and the paired difference is ${+}0.138$ $[{+}0.10,{+}0.18]$, so the retention gap
is not attributable to sampling. On the held-out set the final-round gap is small but
consistently positive---with $\mathcal{R}$ higher at every round (paired difference
${+}0.033$ $[{-}0.04,{+}0.10]$ at $t{=}4$, spanning zero)---consistent with a cumulative
trend rather than a per-round significant difference. On the evolution set the two
conditions track each other at every round (paired difference ${+}0.010$
$[{-}0.06,{+}0.08]$ at $t{=}4$, spanning zero), so retaining historical cases during evolution does
not measurably limit learning from the current round.

\subsection{Robustness to pipeline stochasticity}
\label{app:robustness}
The sweep in Table~\ref{tab:ablation-default} is built to isolate
hyperparameters, but it doubles as a repeated-run study. 
Every stage of the loop is stochastic (diagnosis, consolidation, skill rewriting,
and the agent runs that score each candidate),
so each row is an independent end-to-end execution that re-rolls all of it.
Each also runs under a handicap: a configuration deliberately moved off the default,
and a smaller evolution set, so sparser signal to improve skills from.
Although none of these settings matches the default, their totals average pass@$k$---outside the two
intermediate batch sizes---stay close to one another and match or exceed that of the
strongest baseline, which was evolved on the full set.
That exception is the sensitivity S1 characterizes, not a random excursion:
outside it, quality holds under both configuration and execution randomness.

\subsection{Cost stability and deployment overhead}
\label{app:cost-stability}
Read as repeated runs (\S\ref{app:robustness}), the same sweep settles what the reported
costs are worth. Cost is reported in tokens throughout: billing is a fixed rate per token
and output tokens are generated one at a time, so monetary cost is linear in the counts
below, and per-call latency close to linear in the output count.

\paragraph{Inference cost.} This is paid on every case for as long as the skills are
deployed. Across the ten runs it holds a narrow band---$3.90$--$4.52$k output tokens per
case, and $343$--$383$k input apart from the two extreme batch sizes, which move input to
$402$k at $|\mathcal{B}_i|{=}1$ and $453$k at $40$: at either extreme the edits come out
unusually fine-grained or unusually broad, and both leave the agent extra work to apply
(\S\ref{sec:ablation}). \textbf{B5} spends $6.48$k output tokens per case, above every run
in that band, so its gap to ESCROW is wider than the spread our own runs produce.

\paragraph{Evolution cost.} This is paid once when the skill bank is built, and does not
recur on the cases the deployed agent serves. Holding batch size at its default, the six
runs that vary selection, consensus, and tolerance spend $78$--$83$k input and $21$--$25$k
output tokens per iteration. Batch size is the one setting that moves this
figure---from $379$k input per iteration at $|\mathcal{B}_i|{=}1$ down to $36$k at $40$---
but the setting cheapest to evolve is also the most expensive to serve
(\S\ref{sec:ablation}). The default takes the other side of that trade: evolution is paid
once, inference on every case thereafter.

\section{$\tau$-bench experiment}
\label{app:main-results}
\label{app:taubench}

Beyond the internal FinDAS domain, ESCROW is evaluated on two public domains of $\tau$-bench~\citep{yao2024tau}, 
keeping $\tau$-bench's original cases, tool APIs, and environment configuration unchanged, 
with ESCROW run under the same default settings as the FinDAS experiments (\S\ref{sec:setup}, shown in Table~\ref{tab:config}). 
In each domain, the $\tau$-bench policy is converted into skills to form the \textbf{B2} (LLM skills) seed, from which \textbf{B3}, \textbf{B4}, \textbf{B5}, and ESCROW all derive: 
\textbf{B3} (Human skills) is the same seed after expert review and editing, while \textbf{B4}, \textbf{B5}, and ESCROW evolve it automatically. 
The policy agent and user simulator run on Claude Sonnet~4.5 with thinking enabled, 
and every case is manually classified into the three categories (general, hallucination, special) with matching feedback. 
Because $\tau$-bench provides few cases per domain,
the evolution/test split is chosen per domain to keep both sides usable:
the evolution set must cover all three categories with enough feedback for the loop to learn from,
and the held-out set must remain large enough to give reliable per-category estimates.

\begin{table}[t]
\centering
\caption{Baseline comparison on the public $\tau$-bench domains (pass@$k$, $k{=}3$; Claude Sonnet~4.5 agent). \emph{Total} is the case-weighted average across categories; inference cost is in thousands of tokens ($k$), input\,\textbar\,output.}
\label{tab:taubench}
\footnotesize
\renewcommand{\arraystretch}{0.95}
\setlength{\tabcolsep}{2.1pt}

\begin{subtable}[t]{0.49\linewidth}
\centering
\resizebox{\linewidth}{!}{%
\begin{tabular}{@{}l cccc c@{}}
\toprule
& \multicolumn{4}{c}{Quality (avg.\ pass@$k$)} & Inf.\ Cost \\
\cmidrule(lr){2-5}\cmidrule(lr){6-6}
Method & Gen$\uparrow$ & Hall$\uparrow$ & Spec$\uparrow$ & Total$\uparrow$ & k tok / case$\downarrow$ \\
\midrule
B1\; Raw LLM            & 0.90 & 1.00 & 0.67 & 0.88 & 80.90 \textbar\ 1.52 \\
B2\; LLM skills         & 0.97 & 0.80 & 0.83 & 0.93 & 112.03 \textbar\ 2.93 \\
B3\; Human skills       & 0.97 & 0.80 & 1.00 & 0.95 & 155.55 \textbar\ 3.01 \\
B4\; Batch update       & 0.97 & 1.00 & 1.00 & 0.98 & 123.59 \textbar\ 2.88 \\
B5\; AutoSkill          & 0.97 & 1.00 & 0.83 & 0.95 & 109.16 \textbar\ 2.69 \\
\textbf{ESCROW (ours)} & \textbf{1.00} & \textbf{1.00} & \textbf{0.83} & \textbf{0.98} & \textbf{109.68 \textbar\ 2.14} \\
\bottomrule
\end{tabular}}
\caption{Baseline comparison on the public $\tau$-bench \textbf{retail} domain.}
\label{tab:taubench-retail}
\end{subtable}%
\hfill
\begin{subtable}[t]{0.49\linewidth}
\centering
\resizebox{\linewidth}{!}{%
\begin{tabular}{@{}l cccc c@{}}
\toprule
& \multicolumn{4}{c}{Quality (avg.\ pass@$k$)} & Inf.\ Cost \\
\cmidrule(lr){2-5}\cmidrule(lr){6-6}
Method & Gen$\uparrow$ & Hall$\uparrow$ & Spec$\uparrow$ & Total$\uparrow$ & k tok / case$\downarrow$ \\
\midrule
B1\; Raw LLM            & 1.00 & 0.71 & 0.67 & 0.84 & 93.69 \textbar\ 1.82 \\
B2\; LLM skills         & 1.00 & 0.71 & 0.83 & 0.88 & 129.74 \textbar\ 3.51 \\
B3\; Human skills       & 0.92 & 0.71 & 0.67 & 0.80 & 180.15 \textbar\ 3.61 \\
B4\; Batch update       & 1.00 & 0.86 & 0.67 & 0.88 & 143.13 \textbar\ 3.45 \\
B5\; AutoSkill          & 1.00 & 0.86 & 0.83 & 0.92 & 126.42 \textbar\ 3.23 \\
\textbf{ESCROW (ours)} & \textbf{1.00} & \textbf{1.00} & \textbf{0.83} & \textbf{0.96} & \textbf{127.02 \textbar\ 2.57} \\
\bottomrule
\end{tabular}}
\caption{Baseline comparison on the public $\tau$-bench \textbf{airline} domain.}
\label{tab:taubench-airline}
\end{subtable}
\end{table}

\subsection{Retail domain}
\label{app:taubench-retail}

\paragraph{Setting.} This domain provides $110$ cases, which we manually classify by their outcome into $78$ general, $15$ hallucination, and $17$ special.
A $65/35$ evolution/test split then leaves $70$ cases for evolution ($49$ general, $10$ hallucination, $11$ special) 
and a $40$-case held-out test set ($29$ general, $5$ hallucination, $6$ special).

\paragraph{Results.} Table~\ref{tab:taubench-retail} reports average pass@$k$ on this held-out test set.
The public $\tau$-bench retail policy is compact and already well covered by the seed,
so scores are high across the board and the headroom is small---a near-saturated regime that leaves little room to separate methods on accuracy alone. 
Even here, ESCROW comes out ahead: it ties for the highest overall quality ($0.98$),
is the only method to handle the general category perfectly ($1.00$), and matches the highest on hallucination ($1.00$);
on special, where scores rest on few cases and move in coarse steps, it remains competitive ($0.83$).
What sets ESCROW apart is reaching this at no cost penalty---among the skill-bearing methods, it posts the lowest cost and sits at the high-quality end of the cost--quality frontier,
while the raw LLM (B1) is cheaper only by attaching no skill and is the weakest overall.
The pattern echoes FinDAS---top-tier accuracy at no cost penalty---indicating the approach generalizes beyond our internal domain.

\subsection{Airline domain}


\paragraph{Setting.} Airline provides only $50$ cases in total—the smallest of our three domains—so we adopt an even $50/50$ evolution/test split rather than the $65/35$ used for retail. This keeps enough cases in every category on both sides for a meaningful per-category estimate. The $50$ cases comprise $24$ general, $14$ hallucination, and $12$ special. The split yields $25$ cases for evolution ($12$ general, $7$ hallucination, $6$ special) and a $25$-case held-out test set ($12$ general, $7$ hallucination, $6$ special).

\paragraph{Results.} Table~\ref{tab:taubench-airline} reports average pass@k on the held-out test set. ESCROW achieves the highest overall quality ($0.96$) among all methods. On the hallucination subset, which is the primary target of our approach, ESCROW attains $1.00$, whereas the strongest baseline reaches only $0.86$ (\textbf{B4} and \textbf{B5}). On general cases, ESCROW remains at the ceiling shared by the strongest automatic methods ($1.00$), while on special cases it achieves a competitive $0.83$. Because the Airline benchmark contains only $50$ total cases, with $7$ held-out hallucination cases and $6$ held-out special cases, these category-level estimates are necessarily coarse and should be interpreted with appropriate caution. Nevertheless, the qualitative pattern is consistent with both FinDAS and Retail: ESCROW remains among the highest-quality methods while avoiding any cost penalty.

\section{Supplementary method details}
\label{app:method-details}

\subsection{Non-regression gate: semantics and scope}
\label{app:gate-semantics}

The per-category admission gate (Eq.~\ref{eq:gate}) serves as the primary mechanism for preventing regression during skill evolution. We detail its semantics below.

\paragraph{What the gate enforces.}
Every candidate child $\mathcal{P}'$ must satisfy $\mathrm{PR}_c(\mathcal{P}') \geq (1-\epsilon)\,\mathrm{PR}_c(\mathcal{P}_0)$ for all categories $c$, evaluated over the current evolution set $\mathcal{F}$ plus a sampled subset of the regression pool $R_{\text{eval}} \subset \mathcal{R}$. 
If no candidate clears the gate in a given iteration, the baseline $\mathcal{P}_0$ is retained unchanged.

\paragraph{Deterministic vs.\ probabilistic coverage.}
The gate provides a \emph{deterministic} guard on the evaluated subset $\mathcal{F} \cup R_{\text{eval}}$: every case in this set has been executed and scored, and the per-category floor is enforced exactly (up to $\epsilon$). For the remainder of $\mathcal{R}$ not sampled in a given iteration, the protection is \emph{probabilistic}---it relies on the cumulative coverage achieved by repeated stratified sampling across iterations (quantified below).

\subsection{Regression-pool sampling strategy}
\label{app:regression-sampling}

\paragraph{Pool composition and per-iteration sample size.}
The regression pool $\mathcal{R}$ accumulates all previously resolved cases across successive runs. At each evolution iteration $t$, we draw a regression sample $R_{\text{eval}}$ sized at $50\%$ of the current-round evaluation set (i.e., proportional to the task size rather than to $|\mathcal{R}|$). This keeps per-iteration evaluation cost stable and predictable as the pool grows. Sampling is stratified by category to ensure minority categories (e.g., \emph{special}) are never under-represented.

\paragraph{Sampling mechanics.}
The evaluation budget $|R_{\text{eval}}|$ is allocated in two layers. \textbf{(i)~Skill-targeted priority:} cases associated with skills modified in the current iteration (identified via the same case-skill associations used for anchor-case retrieval; \S\ref{sec:propose}) are sampled first, directly testing whether the edit introduced a regression on its primary dependents. Cases already supplied as anchor constraints in the rewrite prompt are excluded from the sampling pool, so the gate never credits behavior the rewriter was explicitly conditioned on. \textbf{(ii)~Stratified random:} the remaining budget is filled by category-stratified random sampling from the rest of $\mathcal{R}$, covering potential side-effects where editing one skill indirectly affects cases governed by other skills. The split between the two layers is configurable per deployment. Within each iteration, sampling is without replacement; across iterations, sampling is independent, so the gate re-checks a different cross-section on each round and progressively builds cumulative coverage.

\subsection{Coverage under pool growth}
\label{app:coverage-growth}

A natural concern is that as $\mathcal{R}$ grows across successive runs, a fixed-size sample provides diminishing per-case coverage. We show that, under mild assumptions, the two-layer sampling design (\S\ref{app:regression-sampling}) maintains a high probability of \emph{detecting} a regression---i.e., sampling at least one case that exhibits the failure---independent of pool size.

\paragraph{Assumptions.}

\begin{enumerate}
\item[\textbf{A1}] \textbf{(Skill-level clustering.)} We denote by $\mathrm{skills}(x) \subseteq \mathcal{S}$ the set of skills associated with case $x$ (the same association used for anchor-case retrieval; \S\ref{sec:propose}), and by $\mathcal{R}_j = \{x \in \mathcal{R} \mid s_j \in \mathrm{skills}(x)\}$ the cases whose resolution depends on skill $s_j$. If a rewrite to $s_j$ causes regression, at least a fraction $\alpha > 0$ of cases in $\mathcal{R}_j$ will exhibit that regression. This is a reasonable expectation given that skills are shared procedural abstractions: cases resolved by the same skill share overlapping governing logic, so a regression-inducing edit typically manifests across a non-negligible subset rather than a single isolated case.
\item[\textbf{A2}] \textbf{(Sub-linear skill bank growth.)} The skill bank size $|\mathcal{S}|$ grows over time (as new policies are introduced or evolution creates new skills) but at a rate much lower than the pool: $|\mathcal{S}| = o(|\mathcal{R}|)$. This holds because each skill encodes a general paradigm for solving a class of problems and thus covers many cases, whereas the pool grows with every resolved feedback instance.
\item[\textbf{A3}] \textbf{(Positive skill share.)} For any actively maintained skill $s_j$, its pool fraction $p_j = |\mathcal{R}_j|/|\mathcal{R}|$ remains bounded away from zero as $|\mathcal{R}| \to \infty$; formally, $\liminf_{|\mathcal{R}|\to\infty} p_j > 0$. This does \emph{not} require uniform distribution across skills---it only requires that a skill in active use continues to accumulate resolved cases at a rate that keeps pace with overall pool growth. A2 ensures that the number of skills does not grow fast enough to dilute every skill's share to zero.
\end{enumerate}

\paragraph{Primary defense: Targeted-layer detection.}
The skill-targeted layer (layer~i in \S\ref{app:regression-sampling}) samples directly from $\mathcal{R}_j$ when skill $s_j$ is modified. Under A1, a fraction $\alpha$ of those cases exhibit the regression. If the targeted layer draws $n_t$ cases from $\mathcal{R}_j$, the single-iteration detection probability is:
\begin{equation}\label{eq:detect-targeted}
p_{\text{detect}}^{\text{targeted}}(s_j) \;=\; 1 - (1-\alpha)^{n_t}
\end{equation}
where $n_t = \min(|R_{\text{eval}}^{(i)}|,\, |\mathcal{R}_j|)$ and $|R_{\text{eval}}^{(i)}|$ is the targeted-layer budget. Crucially, this expression contains \emph{neither} $|\mathcal{R}|$ nor $p_j$---it is \textbf{inherently pool-independent}.

\paragraph{Secondary defense: Random-layer detection.}
The stratified-random layer (layer~ii) covers side-effects: when skill $s_k$ is edited, the targeted layer checks cases in $\mathcal{R}_k$, but the edit may indirectly affect cases governed by a \emph{different} skill $s_j$ ($j \neq k$). Let $\beta > 0$ denote the fraction of cases in $\mathcal{R}_j$ that exhibit such a side-effect regression (analogous to A1 for indirect effects; typically $\beta \leq \alpha$ since indirect effects are weaker). Drawing $n_r$ cases from $\mathcal{R}$ (modeled as uniform for a conservative lower bound; stratification only improves minority-category coverage), the probability of sampling at least one affected case from $\mathcal{R}_j$ is:
\begin{equation}\label{eq:detect-random}
p_{\text{detect}}^{\text{random}}(s_j) \;=\; 1 - (1-\beta\, p_j)^{n_r}
\end{equation}

\paragraph{Approximation note.}
Both Eq.~\ref{eq:detect-targeted} and Eq.~\ref{eq:detect-random} use the with-replacement (binomial) model. Since actual sampling is without replacement, the true detection probabilities are at least as high (negative correlation only helps). The approximation is tight when the pool is much larger than the sample ($|\mathcal{R}_j| \gg n_t$ for the targeted layer; $|\mathcal{R}| \gg n_r$ for the random layer), which holds in practice.

\paragraph{Pool-independence of detection.}
Neither Eq.~\ref{eq:detect-targeted} nor Eq.~\ref{eq:detect-random} contains $|\mathcal{R}|$ directly. The targeted layer depends only on $\alpha$ and $n_t$; the random layer depends on $\beta$, $p_j$, and $n_r$. Under A3, $p_j$ remains bounded away from zero, so $\beta\, p_j$ does not vanish as the pool grows. Detection probability therefore does not fall in proportion to $|\mathcal{R}|$.

\paragraph{Contrast with naive per-case coverage.}
Under uniform random sampling without skill structure, the probability that a \emph{specific} case is sampled at least once across $T$ iterations is:
\begin{equation}\label{eq:naive-coverage}
p_{\text{case}} = 1 - \left(1 - \frac{|R_{\text{eval}}|}{|\mathcal{R}|}\right)^T,
\end{equation}
which degrades as $|\mathcal{R}|$ grows. The skill-level analysis avoids this degradation: detecting a regression on $s_j$ requires hitting \emph{any one} of the $\alpha\,|\mathcal{R}_j|$ affected cases (targeted layer) or sampling any case whose regression probability is $\beta\, p_j$ per draw (random layer), rather than a specific individual case.

\paragraph{Rate of degradation.}
Detection can degrade through one channel only: $p_j$ drifting toward zero for a skill still being edited. Under A2, the number of skills grows sub-linearly, so the average share $1/|\mathcal{S}|$ shrinks slowly; A3 further ensures that any actively maintained skill's share stays positive. Thus any degradation is driven by the slow pace of skill-bank expansion---substantially slower than pool growth. For skills that become dormant (no new cases), $p_j$ may shrink, but dormant skills are not edited and therefore do not require regression detection.

\section{Limitations}\label{app:limitation}
\paragraph{Generalization gap.}
Our primary evidence comes from a single financial document-auditing system, FinDAS.
The public $\tau$-bench domains show ESCROW is not specific to FinDAS, but they are small,
with the retail domain close to saturated, so they remain preliminary transfer probes and do
not establish how far ESCROW extends; broader domain evidence is future work.
The gap is partly structural: no public benchmark tests what ESCROW targets, and the closest option, document QA, asks fixed questions over documents without exercising skill execution (\S\ref{sec:setup}).

\paragraph{Dependence on feedback quality.}
ESCROW learns from operational feedback, so its gains scale with feedback quality---larger
under accurate, domain-specific feedback, smaller under sparse or noisy signal (e.g.,
$\tau$-bench retail). Consensus mutation and the per-category gate discard unreliable edits,
which limits how much a noisy signal degrades the deployed skills but admits fewer edits, so
improvement is slower.

\paragraph{No model fine-tuning.}
Because ESCROW freezes the model and edits only skills, it cannot add capabilities the base
model lacks. This is deliberate---it avoids the training cost and labelled data fine-tuning
needs, and keeps every change an auditable natural-language diff---but it bounds what
maintenance alone can achieve.

\paragraph{No code or data release.}
FinDAS runs on proprietary policies and real customer documents; our institution has
approved disclosure of the method and its measurements only. We instead report every
cold-start and evolution prompt, all hyperparameters, and the full protocol in the
appendix, and run the same pipeline on public $\tau$-bench, so the claims can be examined
without proprietary data.

\paragraph{Evaluation dominates loop cost.} 
Re-scoring every case through full agent runs makes evaluation one to two orders of magnitude
costlier than all other modules combined, currently capping the loop at a few hundred cases.
This cost buys the non-regression guard---real re-scoring, not heuristics---and is paid during
evolution, never at deployment; cheaper surrogate evaluators are the natural relaxation.

\newpage
\section*{NeurIPS Paper Checklist}

\begin{enumerate}

\item {\bf Claims}
    \item[] Question: Do the main claims made in the abstract and introduction accurately reflect the paper's contributions and scope?
    \item[] Answer: \answerYes{}
    \item[] Justification: The introduction's contributions are each realized in the paper: the framing of continuous improvement as guarded, continual maintenance of enterprise agents (\S\ref{sec:problem}); the mechanisms that implement it (\S\ref{sec:retrieval}--\S\ref{sec:frontier}), each separately ablated in Table~\ref{tab:ablation-default}; and the deployment properties---cold-starting from policies alone and emitting a reviewable diff per change (\S\ref{sec:cold_start}). The accuracy--cost claim is measured against five baselines on the production FinDAS system (Table~\ref{tab:main}), and we claim generality as far as we measure it: the same profile holds on two public $\tau$-bench domains (Table~\ref{tab:taubench}), evidence that the method carries beyond FinDAS.

\item {\bf Limitations}
    \item[] Question: Does the paper discuss the limitations of the work performed by the authors?
    \item[] Answer: \answerYes{}
    \item[] Justification: The Limitations section (\S\ref{app:limitation}) discusses the generalization gap from evidence on a single production system, dependence on operational feedback quality, the ceiling imposed by freezing the base model, and the dominance of evaluation cost in the maintenance loop.

\item {\bf Theory assumptions and proofs}
    \item[] Question: For each theoretical result, does the paper provide the full set of assumptions and a complete (and correct) proof?
    \item[] Answer: \answerYes{}
    \item[] Justification: The paper's one analytical result---that the regression-detection probability of our two-layer sampling design does not degrade as the regression pool grows---is given in full in \S\ref{app:coverage-growth}: assumptions \textbf{A1}--\textbf{A3} are stated explicitly and individually justified, Equations~\ref{eq:detect-targeted} and~\ref{eq:detect-random} are derived from them, and the accompanying approximation note records that the with-replacement model used makes both a conservative lower bound on the true detection probability. Elsewhere in the paper, Equations~\ref{eq:objective}, \ref{eq:gate} and \ref{eq:dominance} define the optimization objective, the admission gate and the dominance relation the system implements; they are definitions rather than results requiring proof.

    \item {\bf Experimental result reproducibility}
    \item[] Question: Does the paper fully disclose all the information needed to reproduce the main experimental results of the paper to the extent that it affects the main claims and/or conclusions of the paper (regardless of whether the code and data are provided or not)?
    \item[] Answer: \answerYes{}
    \item[] Justification: The full protocol is disclosed: the evaluation metric and scoring procedure (\S\ref{sec:problem} and \S\ref{sec:setup}), the default configuration together with every knob varied in the ablation (Table~\ref{tab:config} and \S\ref{sec:ablation}), the complete text of every module prompt (\S\ref{app:prompts}), the case construction and sampling design (\S\ref{app:dataset} and \S\ref{app:regression-sampling}), and representative skills and tool definitions (\S\ref{app:skill-corpus} and \S\ref{app:tools}). The $\tau$-bench experiment (\S\ref{app:taubench}) runs the same pipeline on a public benchmark and is reproducible.

\item {\bf Open access to data and code}
    \item[] Question: Does the paper provide open access to the data and code, with sufficient instructions to faithfully reproduce the main experimental results, as described in supplemental material?
    \item[] Answer: \answerNo{}
    \item[] Justification: FinDAS runs on proprietary internal policies and real customer financial documents, and our institution has approved disclosure of the method and its measurements only; this is stated explicitly in the Limitations section (\S\ref{app:limitation}). To compensate we report all prompts, hyperparameters and the full protocol in the appendix, and run the identical pipeline on the public $\tau$-bench benchmark.

\item {\bf Experimental setting/details}
    \item[] Question: Does the paper specify all the training and test details (e.g., data splits, hyperparameters, how they were chosen, type of optimizer) necessary to understand the results?
    \item[] Answer: \answerYes{}
    \item[] Justification: \S\ref{sec:setup} gives the evaluation set construction, the three operational categories, the two disjoint $210$-case splits, and the frozen base models---Claude Sonnet~4.6 for every ESCROW module and the FinDAS policy agent, Claude Sonnet~4.5 for the $\tau$-bench policy agent and user simulator (\S\ref{app:taubench}). The unbiased pass@$k$ estimator over $n$ trials per case is defined in \S\ref{sec:problem}; the reported setting is $k{=}3$, stated in the captions of Tables~\ref{tab:main} and~\ref{tab:ablation-default} and in \S\ref{app:significance}. Table~\ref{tab:config} gives the default configuration and \S\ref{sec:ablation} states how each ablated setting deviates from it. No model weights are trained, so there is no optimizer or learning-rate schedule to report.

\item {\bf Experiment statistical significance}
    \item[] Question: Does the paper report error bars suitably and correctly defined or other appropriate information about the statistical significance of the experiments?
    \item[] Answer: \answerYes{}
    \item[] Justification: Every reported total pass@$k$ in Tables~\ref{tab:main} and~\ref{tab:ablation-default} carries a $95\%$ percentile bootstrap confidence interval from case-level resampling ($B{=}10{,}000$), and the multi-cycle comparison reports paired-difference intervals and states which of them span zero (Table~\ref{tab:regression-ci}). Because marginal intervals overlap, the lead over the strongest baseline is additionally tested as a paired comparison on the same $210$ held-out cases: a paired bootstrap interval excluding zero and McNemar's exact test ($p{=}0.0026$), reported with its null hypothesis and discordant-case counts in \S\ref{app:significance}. \S\ref{app:regression-ci} describes the intervals for the multi-cycle study and \S\ref{app:robustness} reports variation across repeated end-to-end pipeline runs.

\item {\bf Experiments compute resources}
    \item[] Question: For each experiment, does the paper provide sufficient information on the computer resources (type of compute workers, memory, time of execution) needed to reproduce the experiments?
    \item[] Answer: \answerYes{}
    \item[] Justification: No model weights are trained, so no GPU compute is involved: every experiment is API inference against a frozen model. The relevant resource is token consumption, which is reported throughout as inference cost per case and evolution cost per iteration (Tables~\ref{tab:main} and~\ref{tab:ablation-default}), priced in USD in Figure~\ref{fig:admission}, and analysed in \S\ref{app:cost-stability}.
    
\item {\bf Code of ethics}
    \item[] Question: Does the research conducted in the paper conform, in every respect, with the NeurIPS Code of Ethics \url{https://neurips.cc/public/EthicsGuidelines}?
    \item[] Answer: \answerYes{}
    \item[] Justification: The research conforms with the NeurIPS Code of Ethics. It employs no crowdworkers and releases no data or models. The FinDAS documents it operates on are internal and were anonymized and handled throughout under our institution's existing privacy protocols for customer financial data; the public $\tau$-bench domains are used unchanged under their published terms. To preserve anonymity, the submission does not name the institution operating FinDAS or its internal review procedures.

\item {\bf Broader impacts}
    \item[] Question: Does the paper discuss both potential positive societal impacts and negative societal impacts of the work performed?
    \item[] Answer: \answerYes{}
    \item[] Justification: The Conclusion (\S\ref{sec:conclusion}) discusses both directions. On the negative side it acknowledges the potential for algorithmic bias in automated enterprise workflow. On the positive side, ESCROW is designed to reduce that risk and to preserve transparency and human accountability: admission is settled by evidence rather than by an LLM's judgement---no category may fall below its floor and previously resolved cases are retested at every update (\S\ref{sec:gate})---and every behavioural change reaches production as a versioned, human-readable natural-language diff that a compliance officer can review and approve. It also states that a production deployment would still require rigorous, ongoing audits for demographic bias.
    
\item {\bf Safeguards}
    \item[] Question: Does the paper describe safeguards that have been put in place for responsible release of data or models that have a high risk for misuse (e.g., pre-trained language models, image generators, or scraped datasets)?
    \item[] Answer: \answerNA{}
    \item[] Justification: The paper releases no data, models or scraped assets, so there is nothing with a misuse risk to safeguard.

\item {\bf Licenses for existing assets}
    \item[] Question: Are the creators or original owners of assets (e.g., code, data, models), used in the paper, properly credited and are the license and terms of use explicitly mentioned and properly respected?
    \item[] Answer: \answerYes{}
    \item[] Justification: The public benchmark ($\tau$-bench~\citep{yao2024tau}, used with its original cases, tool APIs and environment configuration unchanged), the agent runtime (DeepAgents~\citep{deepagents2024}) and the base models (Claude Sonnet~4.6~\citep{claude-sonnet-4-6} for the FinDAS experiments and Claude Sonnet~4.5~\citep{claude-sonnet-4-5} for the $\tau$-bench agent and user simulator) are cited at their point of use in \S\ref{sec:setup}, Table~\ref{tab:config} and \S\ref{app:taubench}, with the version of each stated, and are used within their published terms of use.

\item {\bf New assets}
    \item[] Question: Are new assets introduced in the paper well documented and is the documentation provided alongside the assets?
    \item[] Answer: \answerNA{}
    \item[] Justification: No new assets are released with the paper. The FinDAS skills and cases remain internal, for the reasons given in the Limitations section (\S\ref{app:limitation}).

\item {\bf Crowdsourcing and research with human subjects}
    \item[] Question: For crowdsourcing experiments and research with human subjects, does the paper include the full text of instructions given to participants and screenshots, if applicable, as well as details about compensation (if any)? 
    \item[] Answer: \answerNA{}
    \item[] Justification: The paper involves no crowdsourcing and no research with human subjects. The operational feedback it consumes is drawn from logs that internal domain reviewers produce in the ordinary course of their work, not from a study run for this paper.

\item {\bf Institutional review board (IRB) approvals or equivalent for research with human subjects}
    \item[] Question: Does the paper describe potential risks incurred by study participants, whether such risks were disclosed to the subjects, and whether Institutional Review Board (IRB) approvals (or an equivalent approval/review based on the requirements of your country or institution) were obtained?
    \item[] Answer: \answerNA{}
    \item[] Justification: No research with human subjects was conducted, so no IRB or equivalent review was required and no participants were exposed to risk. The operational feedback consumed by the method comes from logs that internal domain reviewers produce in the ordinary course of their work, and the document data is anonymized and handled under our institution's existing privacy protocols for customer financial data. In keeping with the anonymity requirement, we do not name the institution or its review procedures.

\item {\bf Declaration of LLM usage}
    \item[] Question: Does the paper describe the usage of LLMs if it is an important, original, or non-standard component of the core methods in this research? Note that if the LLM is used only for writing, editing, or formatting purposes and does \emph{not} impact the core methodology, scientific rigor, or originality of the research, declaration is not required.
    \item[] Answer: \answerYes{}
    \item[] Justification: LLMs are the core object of study, not a writing aid: they act as the agent under maintenance and as the diagnoser, consolidator, rewriter and evaluator modules of ESCROW. \S\ref{sec:cold_start}--\S\ref{sec:frontier} describe each module, Table~\ref{tab:config} names the models used, and \S\ref{app:prompts} gives the full prompt for every one of them.

\end{enumerate}

\end{document}